\documentclass{article}

\usepackage{iclr2027_conference,times}
\iclrfinalcopy

\usepackage[utf8]{inputenc}
\usepackage[T1]{fontenc}
\usepackage[hypertexnames=false]{hyperref}
\usepackage{url}
\usepackage{booktabs}
\usepackage{array}
\usepackage{multirow}
\usepackage{amsfonts}
\usepackage{amsmath}
\usepackage{amssymb}
\usepackage{nicefrac}
\usepackage{microtype}
\usepackage{xcolor}
\usepackage{colortbl}
\usepackage{graphicx}
\usepackage{float}
\usepackage{capt-of}
\usepackage{placeins}

\definecolor{bestshade}{RGB}{255,220,220}
\definecolor{secondshade}{RGB}{255,244,190}
\newcommand{\bestnum}[1]{\begingroup\setlength{\fboxsep}{0pt}\colorbox{bestshade}{#1}\endgroup}
\newcommand{\secondnum}[1]{\begingroup\setlength{\fboxsep}{0pt}\colorbox{secondshade}{#1}\endgroup}
\newenvironment{nolinenumbers}{}{}

\makeatletter
\def\@maketitle{\vbox{\hsize\textwidth
{\LARGE\sc \@title\par}
\lhead{Preprint}
\begin{tabular}[t]{l}\bf\rule{\z@}{16pt}\@author\end{tabular}%
\vskip 0.3in minus 0.1in}}
\makeatother

\title{JointEdit3D: Feed-Forward 3D Scene Editing in a Unified Latent Space}

\author{%
Xinnan Zhu$^{1,2\ast}$ \quad
Ruijie Xu$^{1,2\ast}$ \quad
Jiayu Ying$^{1}$ \quad
Daoguo Dong$^{3}$ \quad
Jiachen Xu$^{4}$ \\
\textbf{Yuan Xie$^{1}$ \quad
Xin Tan$^{1,2\dagger}$} \\[2pt]
$^1$East China Normal University \quad
$^2$Shanghai AI Laboratory \quad
$^3$Fudan University \quad
$^4$Tencent \\[5pt]
Project page: \url{https://xinnan-zhu.github.io/JointEdit3D-Page/}
}

\begin{document}

\maketitle
\vspace{-16pt}

{\let\thefootnote\relax
\footnotetext{$^\ast$Equal contribution. \quad $^\dagger$Corresponding author.}}

\begin{abstract}
Existing 3D scene editing methods typically rely on per-scene optimization over explicit 3D representations or cascaded edit-and-reconstruct pipelines, resulting in high test-time cost, limited 3D awareness, and structural inconsistencies. To couple appearance synthesis with geometry prediction, we adapt a pretrained unified RGB-geometry latent space to feed-forward scene editing. Given a source video and an edited reference image, \textbf{JointEdit3D} performs asymmetric latent inpainting: it observes only the edited RGB reference latent and jointly generates the remaining RGB latents and the entire geometry latent along the source trajectory. JointEdit3D introduces a dedicated SceneAnchor Branch to inject source-scene structure without forcing direct copying, and adopts edit/background-aware losses to balance edited-region fidelity with unedited-content preservation. To address the lack of paired resources for standardized 3D scene editing evaluation, we introduce \textbf{SceneEdit3D-15K}, a dataset with 15K paired editing samples and renderer-provided 3D annotations, together with \textbf{SceneEdit3D-Bench}, a curated 100-sample benchmark. Experiments show that JointEdit3D improves edited-region quality and 3D structural completeness over prior baselines while maintaining competitive background preservation.

\end{abstract}

\section{Introduction}
\label{sec:intro}

Editing a 3D scene requires modifying its appearance and geometry consistently across views while preserving unaffected scene content. Moving an object, for example, means placing it at a new location and reconstructing the surfaces revealed at its original position. We study this problem in a feed-forward setting: given a source video and an edited reference image, we jointly predict edited appearance and geometry along the source trajectory.

Existing methods often rely on 2D editing guidance or separate view editing from 3D reconstruction. Optimization-based methods edit NeRFs~\citep{mildenhall2020nerf} or 3D Gaussian Splatting~\citep{kerbl3Dgaussians} with guidance from pretrained 2D diffusion models~\citep{instructnerf2nerf, gaussianeditor, gaussctrl, editsplat}. Although the representation is 3D, editing guidance is derived from 2D edits of rendered views. These methods often require precise masks, and their per-scene optimization is costly. Edits can also affect regions beyond the specified mask, for example through changes in shadows and reflections. Recent feed-forward methods reduce test-time cost through multi-view inpainting, video propagation followed by reconstruction, or edited-view alignment before a 3D stage~\citep{mvinpainter, editcast3d, omni3dedit, edit3r}. Although these methods can incorporate geometric cues or 3D priors, inconsistencies introduced during view editing can carry into the 3D output.

A natural way to reduce this separation is to edit in a latent space that represents 3D structure. Prior work studies voxel-structured latents for object editing~\citep{trellis, fuse3d, voxhammer} and geometry-aware latents for scene generation and reconstruction~\citep{prometheus, gen3c, geometryforcing, gen3r}, but these methods are not designed for reference-guided scene editing. Our key idea is to let edit decisions, appearance changes, and geometry updates share the same latent state. We therefore adapt a pretrained unified RGB-geometry latent space~\citep{gen3r} to 3D scene editing and propose \textbf{JointEdit3D}, which jointly generates edited appearance and geometry rather than editing views first and reconstructing afterward. Given a source video, a single edited reference frame, and an optional language instruction, JointEdit3D predicts a complete edited RGB-geometry latent and decodes it into RGB video and a scene point cloud along the source trajectory. We formulate this as asymmetric RGB-geometry latent inpainting: the edited RGB reference provides the observed target content, while the model jointly generates the remaining RGB latents and the entire geometry latent; the reference provides explicit appearance, placement, and local context for propagating the edit across views.

The source video provides reconstruction evidence but conflicts with the edited reference in regions that should change. We introduce a \emph{SceneAnchor Branch} that receives edit cues from the evolving main-branch state and injects source RGB features into the frozen generator through interleaved residual conditioning, treating the source as an edit-aware preservation prior rather than a target to copy. A standard MSE objective also averages over all latent positions, so sparse edits are underweighted relative to unchanged background; we introduce a set of edit-aware losses that separate edited and background regions in both RGB and geometry latents, while emphasizing positions with larger latent changes and frames farther from the reference.

Progress in 3D scene editing also requires paired supervision and consistent evaluation protocols. Existing 3D editing resources rarely provide paired before/after scene edits with 3D annotations, and evaluations are often limited to 2D metrics or single-operation settings. To address this, we construct \textbf{SceneEdit3D-15K}, a dataset of 15K paired scene edits with RGB videos and renderer-provided 3D annotations, and further curate \textbf{SceneEdit3D-Bench}, a 100-sample held-out benchmark evaluating edit fidelity, background preservation, and 3D structure.

Our contributions are summarized as follows:
\begin{itemize}
    \item We formulate reference-guided 3D scene editing as asymmetric inpainting in a pretrained unified RGB-geometry latent space, jointly predicting edited appearance and geometry along the source trajectory.

    \item We introduce JointEdit3D, combining a SceneAnchor Branch conditioned on the evolving edit state with edit-aware losses to propagate the edit specified by a single reference across views while preserving unedited content, without edit-region masks at inference.

    \item We introduce \textbf{SceneEdit3D-15K}, a dataset of 15K paired scene edits with renderer-provided 3D annotations, and \textbf{SceneEdit3D-Bench} for evaluating edit fidelity, background preservation, and 3D structure.
\end{itemize}

\section{Related Work}
\label{sec:related}

\paragraph{3D scene editing.}
Existing methods differ in how they connect edit guidance with 3D representations. Optimization-based methods update NeRFs or 3D Gaussians per scene with text, image, mask, or segmentation guidance~\citep{spinnerf, instructnerf2nerf, dreameditor, gaussianeditor}, often adding grouping, multi-view consistency, or control mechanisms~\citep{gaussiangrouping, gscream, gaussctrl, dge, editsplat, tipeditor, 3ditscene}. Recent feed-forward methods differ in their conditioning inputs and how edited views are converted into 3D outputs. MVInpainter inpaints masked multi-view RGB~\citep{mvinpainter}. EditCast3D propagates a single-frame 2D edit through video and then reconstructs 3D~\citep{editcast3d}. Omni-3DEdit edits in 2D diffusion latents before decoding 3D~\citep{omni3dedit}. Edit3r applies a 2D editor to the input views and then predicts edited 3D Gaussians in a feed-forward pass~\citep{edit3r}. JointEdit3D instead edits directly in a unified RGB-geometry latent, jointly generating aligned RGB and geometry while preserving unedited source content.

\paragraph{3D latent editing and geometry-aware generation.}
At the object level, prior work explores controllable generation and editing in voxel-structured 3D latent spaces~\citep{trellis, fuse3d, voxhammer}. A related line of scene generation and reconstruction couples synthesis with 3D structure via camera, point-cloud, 3D-cache, or reconstruction features~\citep{seva, director3d, prometheus, gen3c, wvd, geometryforcing, uni3c}, but is not designed for editing existing scenes under an edited reference. Building on Gen3R's pretrained unified RGB-geometry latent space~\citep{gen3r}, our approach introduces reference-guided scene editing with source-scene anchoring to preserve unedited content.

\paragraph{Editing datasets and benchmarks.}
Existing image and video editing datasets provide edit supervision but lack paired 3D annotations~\citep{instructpix2pix, magicbrush, effecterase}. General 3D scene datasets provide real or synthetic data for reconstruction and scene understanding~\citep{scannet, replica, hypersim, threedfront, scannetpp} but are not built as before/after editing corpora. Real-scene inpainting benchmarks such as SPIn-NeRF~\citep{spinnerf} and 360-USID~\citep{aurafusion360} focus on object removal. SceneEdit3D-15K provides large-scale paired scene edits across multiple operations with renderer-provided 3D annotations, while SceneEdit3D-Bench supports standardized evaluation.

\section{Method}
\label{sec:method}

\begin{figure}[htbp]
    \centering
    \setlength{\abovecaptionskip}{3pt}
    \setlength{\belowcaptionskip}{0pt}
    \includegraphics[width=0.90\textwidth]{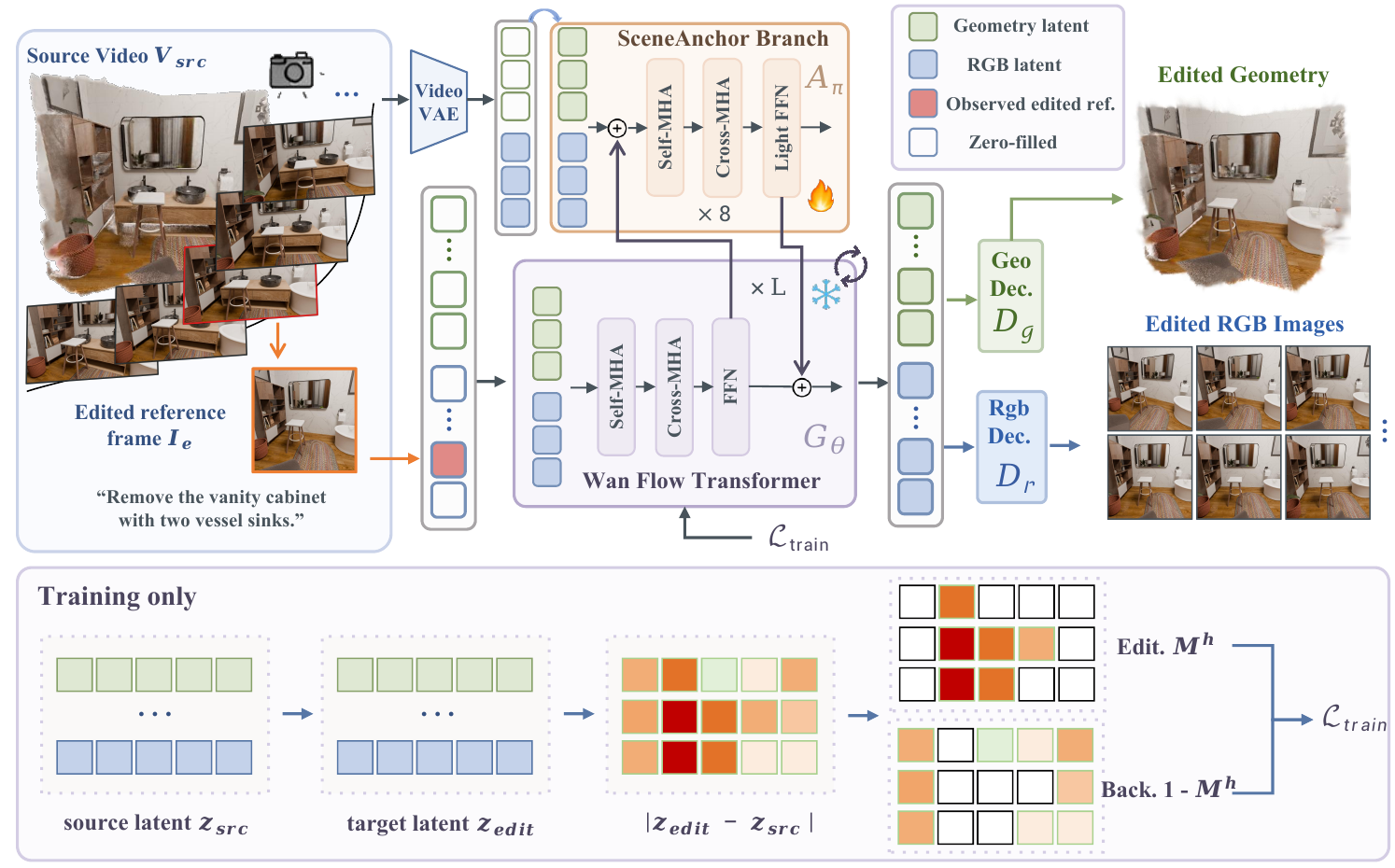}
    \caption{JointEdit3D pipeline: RGB-geometry latent inpainting from a source video and one edited reference frame; source-scene anchoring preserves unedited structure.}
    \label{fig:method}
\end{figure}

\subsection{Preliminaries}
\label{sec:prelim}

We briefly review how Gen3R~\citep{gen3r}, a 3D scene reconstruction model, forms a joint RGB-geometry latent space. Given a multi-view video $\mathcal{V} = \{I^i\}_{i=1}^{N}$, Gen3R encodes appearance and geometry through two aligned streams:
\begin{equation}
\label{eq:rgb_geo_encoding}
\mathbf{z}^{\mathrm{rgb}} = E_{\mathrm{rgb}}(\mathcal{V}), \qquad
\mathbf{h}^{\mathrm{vggt}} = \Phi_{\mathrm{vggt}}(\mathcal{V}), \qquad
\mathbf{z}^{\mathrm{geo}} = A_{\psi}^{\mathrm{enc}}(\mathbf{h}^{\mathrm{vggt}}),
\end{equation}
where $E_{\mathrm{rgb}}$ is the pretrained video VAE encoder~\citep{wan2025}, $\Phi_{\mathrm{vggt}}$ extracts VGGT~\citep{vggt} features, and $A_{\psi}^{\mathrm{enc}}$ is the encoder of the learned Geometry Adapter that maps VGGT tokens into the video latent domain, giving $\mathbf{z}^{\mathrm{rgb}}, \mathbf{z}^{\mathrm{geo}} \in \mathbb{R}^{C \times F \times H \times W}$. Gen3R concatenates the two latents along the spatial width and fine-tunes a Wan-based diffusion backbone to generate the resulting RGB-geometry grid:
\begin{equation}
\label{eq:unified_latent}
\mathbf{z}(\mathcal{V}) = [\,\mathbf{z}^{\mathrm{rgb}};\; \mathbf{z}^{\mathrm{geo}}\,] \in \mathbb{R}^{C \times F \times H \times 2W},
\end{equation}
where the first and second spatial halves store RGB and geometry under matched indices, so generation is joint rather than RGB-only. In JointEdit3D, $G_\theta$ denotes this joint-space Wan backbone post-trained by Gen3R~\citep{gen3r} and $G_{\theta,\pi}$ denotes the full editor after adding the SceneAnchor Branch parameters $\pi$. We reuse the pretrained Geometry Adapter to compute $\mathbf{z}^{\mathrm{geo}}$, and use $\mathbf{z}_{\mathrm{src}}=\mathbf{z}(\mathcal{V}_{\mathrm{src}})$ and $\mathbf{z}_{\mathrm{edit}}=\mathbf{z}(\mathcal{V}_{\mathrm{edit}})$ as the source and edited target latents, respectively, where $\mathcal{V}_{\mathrm{edit}}$ is the edited target video paired with $\mathcal{V}_{\mathrm{src}}$ in our training data. The RGB half of $\mathbf{z}_{\mathrm{edit}}$ comes from the video VAE and its geometry half from the frozen VGGT and Geometry Adapter, so the geometry half of $\mathbf{z}_{\mathrm{edit}}$ is a training pseudo-target and is not provided at inference.

\subsection{Method Overview}
\label{sec:method_overview}

Given a source video $\mathcal{V}_{\mathrm{src}} = \{I_s^i\}_{i=1}^{N}$ and an editing condition $\mathcal{C} = (I_e,\, p)$, comprising an edited reference frame and an optional language instruction, JointEdit3D jointly predicts edited RGB views and geometry along the source trajectory by denoising the unified RGB-geometry latent (Appendix~\ref{app:native_3d_output}). It treats editing as joint latent inpainting rather than RGB-only generation followed by reconstruction, with the edited reference frame $I_e$ specifying the target edit and the source RGB condition providing evidence for preserving unedited content. As shown in Figure~\ref{fig:method}, the framework combines single-frame-guided inpainting (\S\ref{sec:edit_cond}), source-scene preservation through a SceneAnchor Branch (\S\ref{sec:scene_anchor_branch}), and region-decomposed training objectives (\S\ref{sec:training}). We write $\mathbf{ctx} = (\phi_{\mathrm{CLIP}}(I_e),\, \phi_{\mathrm{T5}}(p))$ for the cross-attention context derived from the reference image and instruction.

\subsection{Edit Conditioning via Single-Frame Inpainting}
\label{sec:edit_cond}

In Gen3R reconstruction, every RGB frame is observed and geometry is inferred from it. Editing is asymmetric: only one target RGB frame is observed; the remaining edited RGB views and the entire geometry stream are generated jointly, conditioned on the reference and source video. The source video disagrees with the edited target where changes are required, so it enters through a separate branch (\S\ref{sec:scene_anchor_branch}) rather than as observed target content. We implement this asymmetric observation with latent inpainting: the video VAE encodes $I_e$ alone into $\hat{\mathbf{z}}_e^{\mathrm{rgb}}$, which we place in a zero-filled tensor $\mathbf{y}_{\mathrm{ref}}$ at latent step $k=\lceil r/4\rceil$, where $r$ is the zero-based reference clip index:
\begin{equation}
\label{eq:y_video}
\mathbf{y}_{\mathrm{ref}}[:, k, :, {:W}] = \hat{\mathbf{z}}_e^{\mathrm{rgb}}, \quad \text{all other entries} = 0.
\end{equation}
A four-channel observation mask $\mathbf{m}_{\mathrm{ref}}$ represents the supplied RGB reference under Wan's $4{\times}$ temporal packing; geometry remains unobserved. We concatenate
\begin{equation}
\label{eq:y_concat}
\mathbf{y} = [\,\mathbf{y}_{\mathrm{ref}};\; \mathbf{m}_{\mathrm{ref}}\,] \in \mathbb{R}^{(C+4) \times F \times H \times 2W}
\end{equation}
with the noisy latent $\mathbf{z}_t$ along the channel dimension before the transformer. This observation mask specifies the known frame, not an edit region; temporal grouping and mask packing are detailed in Appendix~\ref{app:implementation_details}.

\subsection{Source-Scene Preservation via SceneAnchor Branch}
\label{sec:scene_anchor_branch}

The edited reference specifies the target change, while the source video provides evidence for preserving unedited scene content. Direct source conditioning must balance preservation against changes specified by the reference, while removing source conditioning weakens background preservation. We therefore use a dedicated \emph{SceneAnchor Branch} $A_\pi$ that receives the evolving denoising state from the frozen main branch and injects source-scene features back as residual corrections. Architecturally, this branch is related to residual adapter and control-branch designs~\citep{controlnet,t2iadapter,ipadapter}, and its role here is source-scene anchoring rather than generic spatial control or image-prompt injection.

\paragraph{Architecture.}
We freeze $G_\theta$ and train only the SceneAnchor Branch parameters $\pi$ to learn residual source-scene adaptation. The branch consumes only the RGB half of the source. Concretely, we feed $\mathbf{c}_{\mathrm{src}} = [\,\mathbf{z}_{\mathrm{src}}^{\mathrm{rgb}};\,\mathbf{0}\,]$, where $\mathbf{z}_{\mathrm{src}}^{\mathrm{rgb}}$ encodes $\mathcal{V}_{\mathrm{src}}$ via the video VAE and the geometry half is zeroed since no source geometry is assumed at inference. The full source latent $\mathbf{z}_{\mathrm{src}}$ from \S\ref{sec:prelim} is computed only during training to derive latent-difference masks and weights (\S\ref{sec:training}). At inference, the conditioning inputs are $\mathbf{c}_{\mathrm{src}}$, $\mathbf{y}$, and $\mathbf{ctx}$; the full source latent $\mathbf{z}_{\mathrm{src}}$ and target latent $\mathbf{z}_{\mathrm{edit}}$ are training-only. A 3D convolutional patch embedding with the same kernel and stride as $G_\theta$ projects this anchor input into an anchor state $\mathbf{c}_0$. The branch follows $G_\theta$'s block structure at hidden width $d_a{=}512$, with learned projections between the branch and backbone widths; each anchor block attaches to a selected backbone layer (Appendix~\ref{app:implementation_details}).

\paragraph{Interleaved residual fusion.}
During denoising, $A_\pi$ is interleaved with selected layers of $G_\theta$, so source conditioning adapts to the current edit state rather than acting as a static copy signal. Let $\mathbf{x}_{\ell_j}$ denote the main-branch hidden tokens at layer $\ell_j$. At the $j$-th injected layer, the current denoising state is projected into the anchor stream and processed by the anchor block:
\begin{equation}
\label{eq:anchor_update}
\mathbf{c}_j \leftarrow \mathrm{AnchorBlock}_j(\mathbf{c}_{j-1} + W_{\mathrm{in}}^{(j)}(\mathbf{x}_{\ell_j}),\; W_e^{(j)}(\mathbf{e}),\; W_{\mathrm{ctx}}^{(j)}(\mathbf{ctx})),
\end{equation}
where $\mathbf{e}$ is the diffusion timestep embedding. The anchor output is mapped back to the main width and added as a residual update:
\begin{equation}
\label{eq:anchor_hint}
\mathbf{x}_{\ell_j} \leftarrow \mathbf{x}_{\ell_j} + W_{\mathrm{out}}^{(j)}(\mathbf{c}_j).
\end{equation}
All $W_{\mathrm{out}}^{(j)}$ are zero-initialized, so the editor reduces to $G_\theta$ at the start of training. Because the anchor stream consumes both the source RGB condition and the evolving main-branch state, its residual updates depend on the current edit hypothesis. This provides a mechanism for balancing source preservation with the requested change without an input edit-region mask. Disabling the feedback degrades both edit-region and background metrics (Table~\ref{tab:ablation}), supporting the use of the evolving main state for source conditioning.

\subsection{Training Objectives}
\label{sec:training}

The base objective follows flow matching~\citep{lipman2023flow} over the edited target latent $\mathbf{z}_{\mathrm{edit}}$. A standard MSE averages over all positions, giving sparse edited regions a small share of the loss weight; we therefore decompose the loss into edit and background regions in each modality. Given noise $\boldsymbol{\epsilon} \sim \mathcal{N}(\mathbf{0}, \mathbf{I})$ and timestep $t \in (0,1)$, we set $\mathbf{z}_t = (1 - t)\mathbf{z}_{\mathrm{edit}} + t\boldsymbol{\epsilon}$. The full editor predicts a velocity $G_{\theta,\pi}(\mathbf{z}_t, t, \mathbf{y}, \mathbf{c}_{\mathrm{src}}, \mathbf{ctx})$ supervised toward $\boldsymbol{\epsilon} - \mathbf{z}_{\mathrm{edit}}$. For each position $i$,
\begin{equation}
\label{eq:per_position_loss}
l_i = \frac{1}{C}\left\|G_{\theta,\pi}(\mathbf{z}_t, t, \mathbf{y}, \mathbf{c}_{\mathrm{src}}, \mathbf{ctx})_i - (\boldsymbol{\epsilon}_i - \mathbf{z}_{\mathrm{edit},i})\right\|_2^2 .
\end{equation}
A training-only temporal weight $a_i$ upweights positions temporally farther from the reference (Appendix~\ref{app:implementation_details}); we write $\tilde{l}_i = a_i l_i$ and apply region weighting separately to the RGB and geometry halves.

\paragraph{Edit-aware region decomposition.}
To prevent the background from dominating the loss, we estimate edited positions from source-target latent changes in each half. For $h \in \{\mathrm{rgb}, \mathrm{geo}\}$ and positions $\Omega_h$,
\begin{equation}
\label{eq:edit_mask}
d_i^{h} =
\frac{\frac{1}{C}\left\|\mathbf{z}_{\mathrm{edit},i}^{h} - \mathbf{z}_{\mathrm{src},i}^{h}\right\|_1}
{\max_{j \in \Omega_h} \frac{1}{C}\left\|\mathbf{z}_{\mathrm{edit},j}^{h} - \mathbf{z}_{\mathrm{src},j}^{h}\right\|_1 + \varepsilon_{\mathrm{den}}},
\qquad
M_i^{h} = \mathbf{1}[d_i^{h} > \tau].
\end{equation}
This produces separate masks for appearance and geometry so the loss can emphasize modality-specific latent changes. These latent-difference masks are used only to construct the training loss; they are not available or required at inference. Appendix~\ref{app:loss_diagnostic} shows training-time masks, inference-time edit responses, and the calibration of $\tau$.

\paragraph{Within-region weighting.}
Inside the edit region, a linear weight $w_i^{h} = 1 + \alpha d_i^{h}$ assigns larger relative loss weights to larger latent changes:
\begin{equation}
\label{eq:region_loss}
\mathcal{L}_{\mathrm{edit}}^{h} =
\frac{\sum_{i \in \Omega_h} w_i^{h} \tilde{l}_i M_i^{h}}{\sum_{i \in \Omega_h} w_i^{h} M_i^{h} + \varepsilon_{\mathrm{den}}},
\quad
\mathcal{L}_{\mathrm{bg}}^{h} =
\frac{\sum_{i \in \Omega_h} \tilde{l}_i (1 - M_i^{h})}{\sum_{i \in \Omega_h} (1 - M_i^{h}) + \varepsilon_{\mathrm{den}}},
\end{equation}
and the two regions are combined per half as $\mathcal{L}_{h} = \lambda\,\mathcal{L}_{\mathrm{edit}}^{h} + (1-\lambda)\,\mathcal{L}_{\mathrm{bg}}^{h}$. The final objective $\mathcal{L}_{\mathrm{train}} = \mathcal{L}_{\mathrm{rgb}} + \gamma_{\mathrm{geo}}\,\mathcal{L}_{\mathrm{geo}}$ balances RGB and geometry-latent supervision, where $\lambda$ controls the edit/background balance and $\gamma_{\mathrm{geo}}$ the relative weight of geometry-latent supervision.

\FloatBarrier

\section{SceneEdit3D-15K Dataset}
\label{sec:dataset}
\label{sec:data_pipeline}
\label{sec:data_generation}
\label{sec:dataset_stats}

Paired editing data with 3D ground truth remain limited. Existing 2D editing datasets~\citep{instructpix2pix, magicbrush} lack paired geometry and multi-view supervision for evaluating scene structure. We introduce \textbf{SceneEdit3D-15K}, a large-scale dataset of 15K paired scene editing samples with language instructions, edited reference frames, edit masks, and renderer-provided 3D annotations, together with \textbf{SceneEdit3D-Bench}, a curated benchmark of 100 held-out samples.

\begin{figure}[t]
    \centering
    \includegraphics[width=0.83\linewidth]{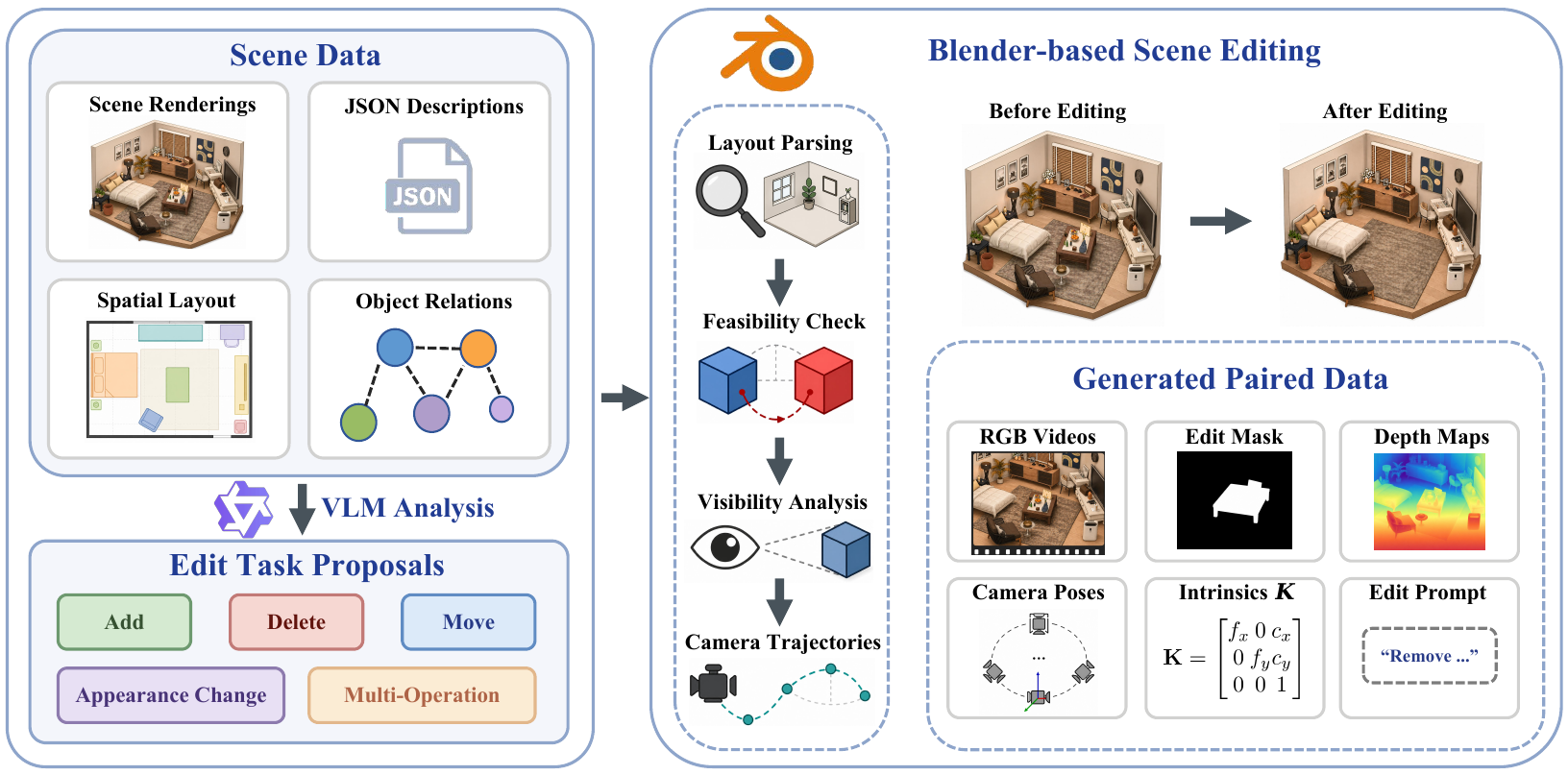}
    \caption{SceneEdit3D-15K data generation pipeline.}
    \label{fig:data_pipeline}
\end{figure}

SceneEdit3D-15K is built from composable indoor scenes in Imaginarium~\citep{imaginarium} whose objects can be individually manipulated in Blender. A VLM analyzes rendered views, object metadata, and layout relations to propose canonical deletion and relocation tasks, while Blender-side procedures generate material/texture appearance changes; Blender then executes each edit deterministically and renders paired source and target videos under identical camera trajectories, with addition and reverse-relocation samples derived by swapping before/after states and reversing frame order. Multi-operation samples combine edits to distinct objects. Training uses the paired RGB videos, one edited reference frame, and a natural-language instruction. Renderer annotations support data preparation, calibration, and evaluation, as detailed in the appendix.

The dataset contains 13,799 training samples from 135 scenes and 1,520 held-out samples from 16 separate scenes. The held-out pool is split at the sample level into 1,420 validation samples and 100 test samples forming SceneEdit3D-Bench, stratified across the five operation types and scene categories, with varied object scales, image locations, and partial occlusions, supporting region-aware evaluation of edited-region quality, background preservation, and 3D structural correctness; generation protocol, quality control, and split details are in Appendix~\ref{app:dataset_details}.

\section{Experiments}
\label{sec:experiments}

\subsection{Experimental Setup}
\label{sec:exp_setup}

\paragraph{Implementation details.}
We instantiate JointEdit3D with a frozen Gen3R-tuned image-camera-conditioned Wan2.1 1.3B DiT backbone $G_\theta$ ($L{=}32$ blocks)~\citep{gen3r}. The SceneAnchor Branch $A_\pi$ is a trainable branch with hidden width $d_a{=}512$ and $K{=}8$ anchor blocks, totaling 70M trainable parameters. We use AdamW with initial learning rate $1{\times}10^{-4}$, bf16 mixed precision, and gradient checkpointing. For the region-decomposed objective: edit-mask threshold $\tau{=}0.15$, within-region weight $\alpha{=}5.0$, temporal weight $\eta{=}0.5$, edit/background balance $\lambda{=}0.5$, and geometry weight $\gamma_{\mathrm{geo}}{=}1.5$. Training runs on the SceneEdit3D-15K split for 10K steps with batch size 10 per device on 8 NVIDIA H200 GPUs. Further implementation details are provided in Appendix~\ref{app:implementation_details}.

\paragraph{Baselines.}
We compare with representative optimization-based and feed-forward methods: SPIn-NeRF~\citep{spinnerf}, Gaussian Grouping~\citep{gaussiangrouping}, GScream~\citep{gscream}, GaussianEditor~\citep{gaussianeditor}, MVInpainter~\citep{mvinpainter}, and Omni-3DEdit~\citep{omni3dedit}, and include SEVA~\citep{seva} as a protocol-mismatched novel-view synthesis diagnostic. Table~\ref{tab:baseline_coverage} summarizes interfaces and operation coverage. A check under Unified weights means one shared set of editor weights covers every supported operation. Baselines use official implementations; input adaptations, comparison scope, and frame budgets are in Appendix~\ref{app:baseline_protocol}.

\begin{table}[tbp]
\centering
\caption{Capability comparison of editing methods (Opt./FF: optimization-based/feed-forward; \(\checkmark\): supported; \(\circ\): protocol-mismatched; --: unsupported under the released interface).}
\label{tab:baseline_coverage}
\begingroup
\scriptsize
\setlength{\tabcolsep}{2.4pt}
\renewcommand{\arraystretch}{0.96}
\newcommand{\cmark}{\(\checkmark\)}
\newcommand{\omark}{\(\circ\)}
\newcolumntype{C}[1]{>{\centering\arraybackslash}m{#1}}
\resizebox{0.98\textwidth}{!}{%
\begin{tabular}{@{}l c *{3}{C{5.8em}} *{5}{C{4.4em}}@{}}
\toprule
\multirow{2}{*}{Method} & \multirow{2}{*}{Type} & \multicolumn{3}{c}{Properties} & \multicolumn{5}{c}{Edit operations} \\
\cmidrule(lr){3-5} \cmidrule(lr){6-10}
& & Mask-free & Pose-free & \shortstack{Unified\\weights} & Delete & Add & Move & Appear. & Multi-op \\
\midrule
SPIn-NeRF~\citep{spinnerf} & Opt. & -- & -- & -- & \cmark & -- & -- & -- & -- \\
Gaussian Grouping~\citep{gaussiangrouping} & Opt. & -- & -- & -- & \cmark & \omark & -- & \omark & -- \\
GScream~\citep{gscream} & Opt. & -- & -- & -- & \cmark & -- & -- & -- & -- \\
GaussianEditor~\citep{gaussianeditor} & Opt. & -- & -- & -- & \cmark & \omark & -- & \omark & -- \\
SEVA~\citep{seva} & FF & \cmark & -- & \cmark & \omark & \omark & \omark & \omark & \omark \\
MVInpainter~\citep{mvinpainter} & FF & -- & \cmark & -- & \cmark & \cmark & \cmark & \cmark & \cmark \\
Omni-3DEdit~\citep{omni3dedit} & FF & \cmark & -- & -- & \cmark & \cmark & \cmark & \cmark & \cmark \\
\textbf{Ours} & FF & \cmark & \cmark & \cmark & \cmark & \cmark & \cmark & \cmark & \cmark \\
\bottomrule
\end{tabular}%
}
\endgroup
\end{table}

\paragraph{Evaluation benchmark and metrics.}
We evaluate on \textbf{SceneEdit3D-Bench} and use 360-USID~\citep{aurafusion360} for real-scene deletion. Trained only on synthetic SceneEdit3D-15K, JointEdit3D also shows qualitative generalization on DAVIS, ScanNet++, CO3D, and casual captures (Appendix~\ref{app:additional_qualitative}). We report PSNR$\uparrow$/LPIPS$\downarrow$ on edit/background regions and 3D metrics against renderer-provided edited geometry (Appendix~\ref{app:3d_metric_protocol}).

\subsection{Main Results}
\label{sec:main_results}

\paragraph{Quantitative comparison.}
We organize comparisons by operation coverage, with \bestnum{best} and \secondnum{second-best} shaded: Table~\ref{tab:delete_results} reports edit-region quality in the shared deletion setting on SceneEdit3D-Bench and 360-USID, and Table~\ref{tab:main_operation_results} reports operation-wise results for methods supporting all five edit types, with separate PSNR$\uparrow$/LPIPS$\downarrow$ on the renderer-defined edit (``Edit'') and background (``Bg.'') regions.

\begin{table}[t]
\centering
\begin{minipage}[c]{0.46\textwidth}
\centering
\caption{Delete edit-region results (PSNR$\uparrow$/LPIPS$\downarrow$).}
\label{tab:delete_results}
\begingroup
\small
\setlength{\tabcolsep}{2.4pt}
\renewcommand{\arraystretch}{1.12}
\newcommand{\delpair}[2]{#1\,/\,#2}
\resizebox{\linewidth}{!}{%
\begin{tabular}{@{}lcc@{}}
\toprule
Method & SceneEdit3D-Bench & 360-USID \\
\midrule
SPIn-NeRF~\citep{spinnerf} & \delpair{19.20}{0.528} & \delpair{15.89}{0.4826} \\
Gaussian Grouping~\citep{gaussiangrouping} & \delpair{20.41}{0.377} & \delpair{16.67}{0.4241} \\
GScream~\citep{gscream} & \delpair{20.38}{0.449} & \delpair{14.37}{0.5725} \\
GaussianEditor~\citep{gaussianeditor} & \delpair{20.26}{0.451} & \delpair{16.23}{0.3975} \\
MVInpainter~\citep{mvinpainter} & \delpair{21.37}{0.406} & \delpair{15.60}{0.5677} \\
Omni-3DEdit~\citep{omni3dedit} & \delpair{24.86}{0.307} & \delpair{17.78}{0.3655} \\
Omni-3DEdit + SceneEdit3D FT & \delpair{\secondnum{28.99}}{\secondnum{0.230}} & \delpair{\secondnum{18.02}}{\secondnum{0.3637}} \\
\textbf{JointEdit3D (Ours)} & \delpair{\bestnum{31.92}}{\bestnum{0.151}} & \delpair{\bestnum{18.57}}{\bestnum{0.3426}} \\
\bottomrule
\end{tabular}
}
\endgroup
\end{minipage}
\hspace{0.08\textwidth}
\begin{minipage}[c]{0.38\textwidth}
\centering
\includegraphics[height=3.2cm]{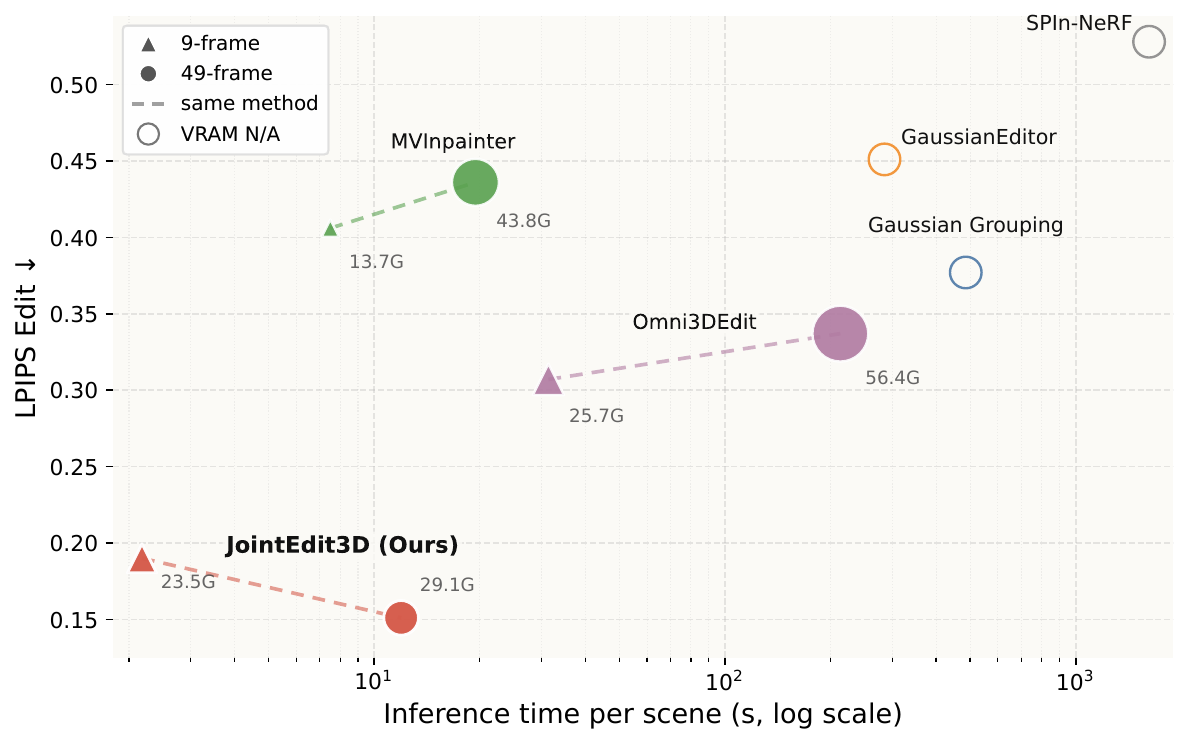}
\captionof{figure}{Efficiency-quality plot.}
\label{fig:runtime_efficiency}
\end{minipage}
\end{table}

\begin{table}[t]
\centering
\caption{Operation-wise PSNR$\uparrow$/LPIPS$\downarrow$ on Bg./Edit regions; Average is the mean over all 100 samples (Appendix~\ref{app:baseline_protocol}).}
\label{tab:main_operation_results}
\begingroup
\setlength{\tabcolsep}{3.2pt}
\renewcommand{\arraystretch}{0.94}
\newcommand{\metricpair}[2]{#1\,/\,#2}
\newcommand{\arealabel}[1]{\multirow{5}{*}{\textbf{#1}}}
\newcolumntype{T}{>{\centering\arraybackslash}p{5.6em}}
\resizebox{\textwidth}{!}{%
\begin{tabular}{@{}>{\centering\arraybackslash}m{2.0em}@{\hspace{6pt}}l TTTTT T@{}}
\toprule
Reg. & Method & Delete & Add & Move & Appearance & Multi-op & Average \\
\midrule
\arealabel{Bg.} & SEVA~\citep{seva} & \metricpair{17.89}{0.421} & \metricpair{17.74}{0.426} & \metricpair{17.69}{0.411} & \metricpair{19.00}{0.378} & \metricpair{17.54}{0.445} & \metricpair{17.93}{0.418} \\
& MVInpainter~\citep{mvinpainter} & \metricpair{\bestnum{33.57}}{\secondnum{0.080}} & \metricpair{\bestnum{33.64}}{\bestnum{0.076}} & \metricpair{\secondnum{30.97}}{\secondnum{0.081}} & \metricpair{\secondnum{33.76}}{\secondnum{0.073}} & \metricpair{\secondnum{30.06}}{\bestnum{0.091}} & \metricpair{\bestnum{32.76}}{\secondnum{0.080}} \\
& Omni-3DEdit~\citep{omni3dedit} & \metricpair{25.00}{0.215} & \metricpair{24.24}{0.217} & \metricpair{24.40}{0.210} & \metricpair{25.88}{0.190} & \metricpair{23.81}{0.235} & \metricpair{24.65}{0.214} \\
& Omni-3DEdit + SceneEdit3D FT & \metricpair{26.43}{0.207} & \metricpair{27.24}{0.195} & \metricpair{26.76}{0.192} & \metricpair{27.07}{0.196} & \metricpair{26.44}{0.218} & \metricpair{26.80}{0.201} \\
& \textbf{JointEdit3D (Ours)} & \metricpair{\secondnum{32.62}}{\bestnum{0.079}} & \metricpair{\secondnum{32.59}}{\secondnum{0.081}} & \metricpair{\bestnum{31.22}}{\bestnum{0.069}} & \metricpair{\bestnum{33.80}}{\bestnum{0.059}} & \metricpair{\bestnum{30.84}}{\secondnum{0.095}} & \metricpair{\secondnum{32.33}}{\bestnum{0.077}} \\
\midrule
\arealabel{Edit} & SEVA & \metricpair{23.39}{0.399} & \metricpair{15.92}{0.523} & \metricpair{17.87}{0.450} & \metricpair{17.50}{0.464} & \metricpair{16.93}{0.496} & \metricpair{18.72}{0.465} \\
& MVInpainter & \metricpair{21.37}{0.406} & \metricpair{20.04}{0.362} & \metricpair{16.99}{0.437} & \metricpair{18.94}{0.349} & \metricpair{17.92}{0.423} & \metricpair{19.55}{0.392} \\
& Omni-3DEdit & \metricpair{24.86}{0.307} & \metricpair{20.63}{0.293} & \metricpair{18.55}{0.377} & \metricpair{20.92}{0.230} & \metricpair{17.03}{0.461} & \metricpair{21.10}{0.323} \\
& Omni-3DEdit + SceneEdit3D FT & \metricpair{\secondnum{28.99}}{\secondnum{0.230}} & \metricpair{\secondnum{23.20}}{\secondnum{0.247}} & \metricpair{\secondnum{22.75}}{\secondnum{0.279}} & \metricpair{\secondnum{24.28}}{\secondnum{0.153}} & \metricpair{\secondnum{21.97}}{\secondnum{0.336}} & \metricpair{\secondnum{24.80}}{\secondnum{0.246}} \\
& \textbf{JointEdit3D (Ours)} & \metricpair{\bestnum{31.92}}{\bestnum{0.151}} & \metricpair{\bestnum{23.88}}{\bestnum{0.195}} & \metricpair{\bestnum{23.19}}{\bestnum{0.244}} & \metricpair{\bestnum{27.67}}{\bestnum{0.115}} & \metricpair{\bestnum{23.77}}{\bestnum{0.263}} & \metricpair{\bestnum{26.63}}{\bestnum{0.187}} \\
\bottomrule
\end{tabular}%
}
\endgroup
\end{table}

Tables~\ref{tab:delete_results} and~\ref{tab:main_operation_results} show that JointEdit3D achieves the strongest edited-region quality on the deletion benchmark and all five SceneEdit3D-Bench operations, with competitive background preservation. Mask-guided baselines receive additional edit-localization information and attain higher background PSNR on some operations. JointEdit3D substantially improves both edit and background metrics over the mask-free Omni-3DEdit. Fine-tuning Omni-3DEdit on SceneEdit3D-15K improves every operation, demonstrating that the dataset benefits another editing architecture. JointEdit3D retains its lead after both models are adapted on the same dataset for 10K updates. Move and Multi-op remain challenging. Gains on real-world 360-USID are smaller than on SceneEdit3D-Bench, likely because its simpler scenes with centrally located target objects leave less room for improvement. Additional object-removal results on Inpaint360GS are in Appendix~\ref{app:inpaint360gs}. Figure~\ref{fig:runtime_efficiency} shows leading edit quality and faster inference under the reported settings.

\begin{figure}[t]
\centering
\includegraphics[width=0.91\textwidth]{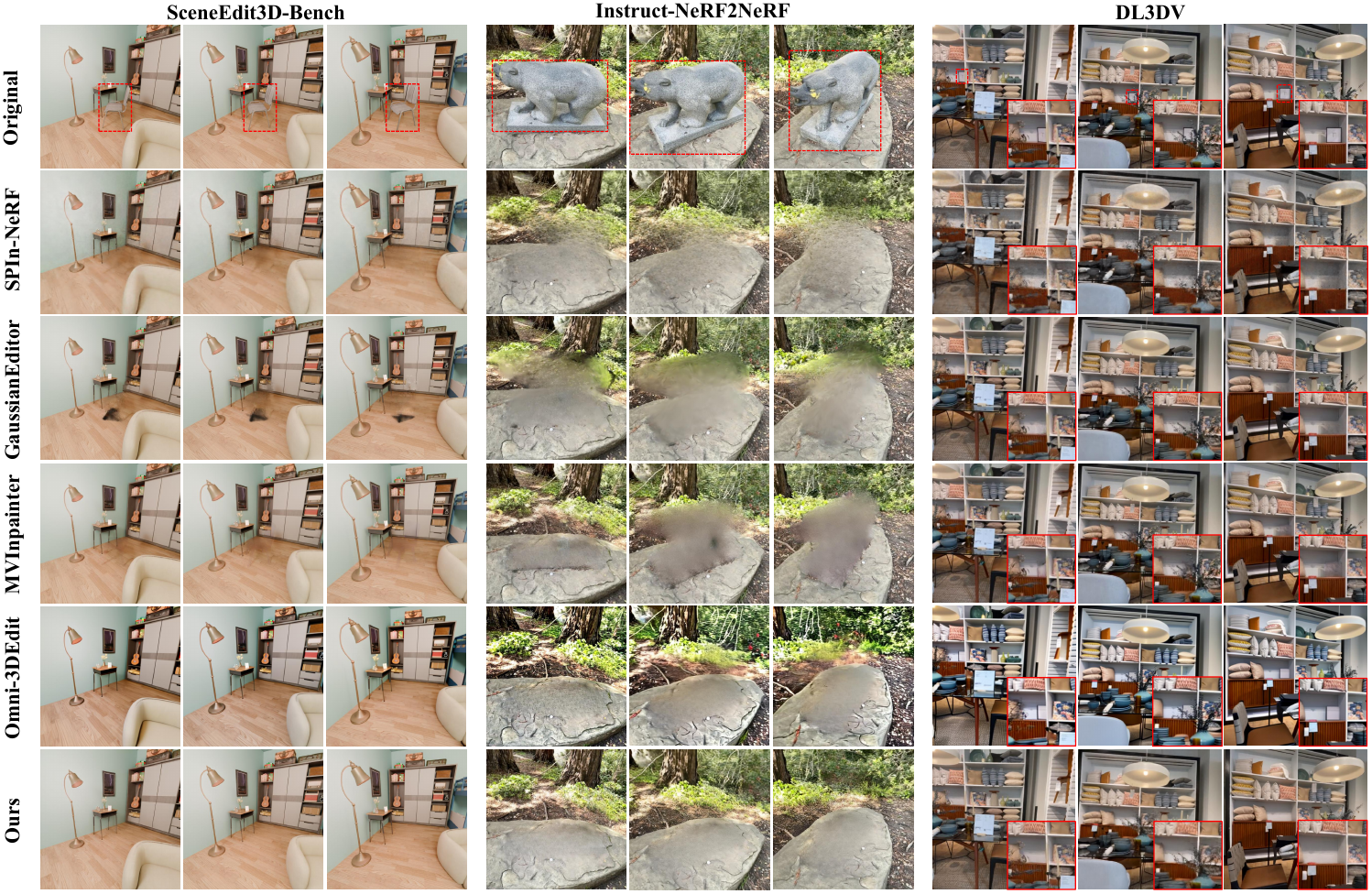}
\caption{Deletion comparison on synthetic SceneEdit3D-Bench and real Instruct-NeRF2NeRF / DL3DV captures~\citep{instructnerf2nerf,dl3dv10k}. Red boxes mark edited regions.}
\label{fig:qualitative_deletion}
\end{figure}

\paragraph{Qualitative deletion results.}
Figure~\ref{fig:qualitative_deletion} compares synthetic SceneEdit3D-Bench and real Instruct-NeRF2NeRF/DL3DV captures~\citep{instructnerf2nerf, dl3dv10k}. SPIn-NeRF and GaussianEditor leave residuals or blur around removed objects. MVInpainter retains boundary discontinuities and shadows, while Omni-3DEdit changes background colors and misses the small photo-frame edit in the cluttered shelf. JointEdit3D removes the targets while preserving nearby structures, including the lamp and shelf contents. Together with the 360-USID metrics, these examples support transfer to real inputs despite synthetic-only training.

\begin{table}[htbp]
\centering
\caption{3D geometry comparison. Confidence filtering uses a GT-adjusted threshold ($^\dagger$; Appendix~\ref{app:region_geometry}). Oracle and filtered rows are diagnostics excluded from ranking.}
\label{tab:external_reconstruction}
\begingroup
\small
\setlength{\tabcolsep}{4pt}
\renewcommand{\arraystretch}{0.94}
\begin{tabular}{@{}l c c c c c@{}}
\toprule
Method & Acc.$\downarrow$ & Comp.$\downarrow$ & F-score$\uparrow$ & Edit CD$\downarrow$ & Bg.\ CD$\downarrow$ \\
\midrule
Target RGB + VGGT (oracle) & 0.0724 & 0.0950 & 0.9619 & 0.1329 & 0.0914 \\
\midrule
SEVA + VGGT & 0.1433 & 0.2244 & 0.8722 & 0.2184 & 0.1825 \\
Omni-3DEdit + VGGT & 0.1165 & 0.1313 & 0.9143 & 0.1878 & 0.1191 \\
JointEdit3D RGB + VGGT & \bestnum{0.0891} & \secondnum{0.1211} & \secondnum{0.9357} & \secondnum{0.1421} & \bestnum{0.0958} \\
\textbf{JointEdit3D (joint output)} & \secondnum{0.1153} & \bestnum{0.0886} & \bestnum{0.9702} & \bestnum{0.1352} & \secondnum{0.0966} \\
\midrule
JointEdit3D, confidence-filtered$^\dagger$ & 0.0872 & 0.1153 & 0.9653 & 0.1382 & 0.0827 \\
\bottomrule
\end{tabular}
\endgroup
\end{table}

\paragraph{3D geometry quality.}
Table~\ref{tab:external_reconstruction} evaluates geometry against the renderer-provided edited cloud using the Gen3R protocol~\citep{gen3r} (Appendix~\ref{app:3d_metric_protocol}). Native joint output and \emph{JointEdit3D RGB + VGGT} share the generated RGB sequence but differ in geometry prediction. Joint output lowers edit-region CD and retains comparable background CD. Without any confidence filtering, the native output achieves better Completeness and F-score than \emph{JointEdit3D RGB + VGGT}. This likely reflects the generative model's ability to infer additional geometry in occluded regions and at object boundaries, where point locations are less certain. Filtering these predictions at a confidence threshold of 0.85 improves Accuracy over the unfiltered output while retaining better Completeness and F-score than the RGB + VGGT reconstruction. This GT-adjusted setting is a post-hoc diagnostic ($^\dagger$; Appendix~\ref{app:region_geometry}). Against Omni-3DEdit + VGGT, both edit-region and background CD improve.

\begin{figure}[t]
\centering
\includegraphics[width=0.96\textwidth,trim=10 18 10 8,clip]{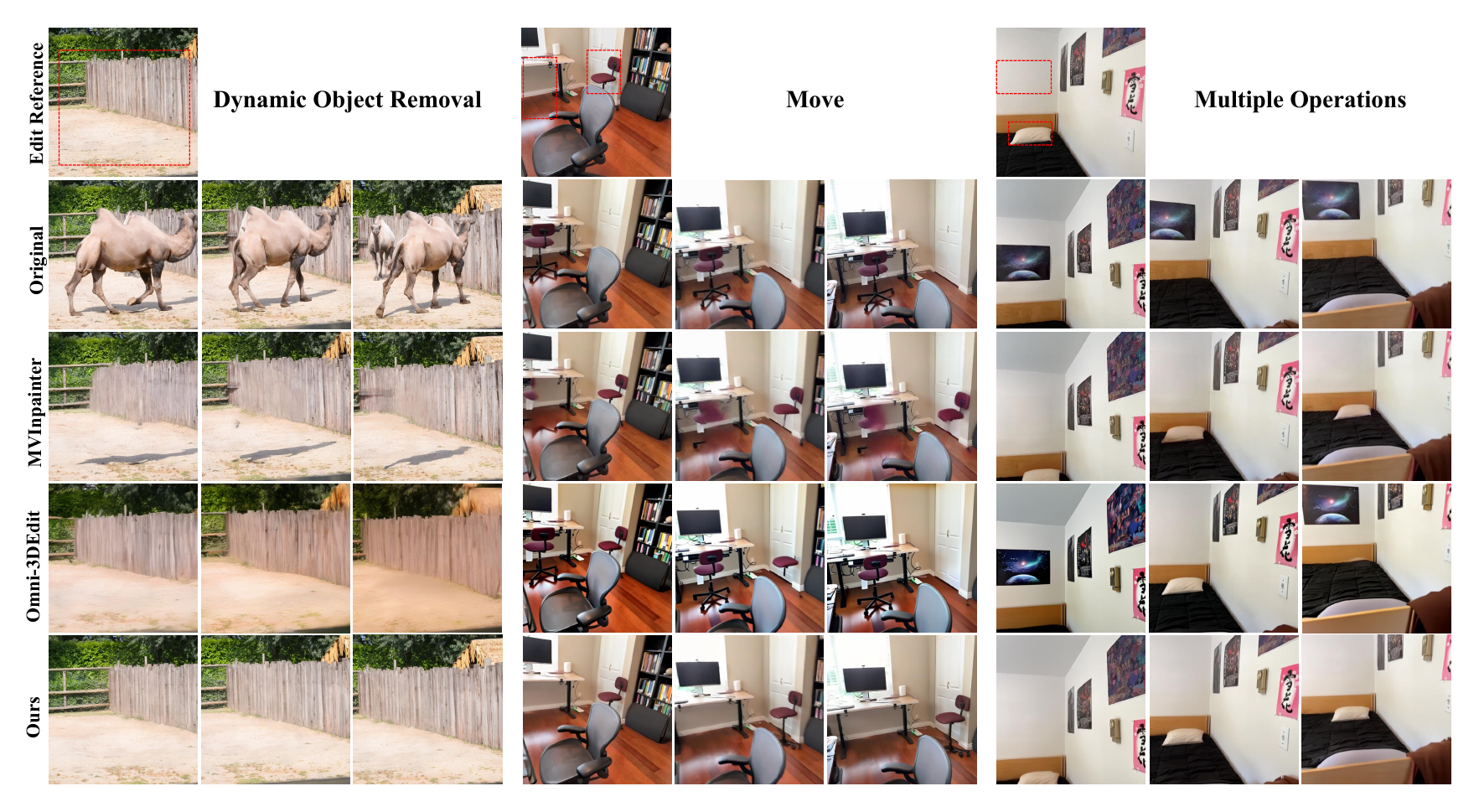}
\caption{Real-scene dynamic object removal, relocation, and multi-operation editing. Red boxes mark edited regions.}
\label{fig:qualitative}
\end{figure}

\paragraph{Real-scene editing across operations.}
Figure~\ref{fig:qualitative} shows dynamic object removal, relocation, and multi-operation editing. MVInpainter leaves the camel's shadow after removal and color discontinuities near multi-edit boundaries. Both baselines struggle with relocation; Omni-3DEdit's moved object appears and disappears across views. JointEdit3D produces cleaner edits while preserving nearby structures across these cases. Appearance editing, occluded removal, and native point-cloud outputs are in Appendix~\ref{app:additional_qualitative}.

\subsection{Ablations and Design Comparisons}
\label{sec:ablation}

Due to computational constraints, all variants in Table~\ref{tab:ablation}, including the full model, use 2K training steps. Appendices~\ref{app:implementation_details} and~\ref{app:main_state_feedback} detail adaptation configurations and the controlled main-state feedback ablation.

\begin{table}[htbp]
\centering
\caption{Component ablations and alternative adaptation designs for JointEdit3D.}
\label{tab:ablation}
\begingroup
\renewcommand{\arraystretch}{0.88}
\resizebox{0.92\textwidth}{!}{%
\begin{tabular}{@{}l cc cc c@{}}
\toprule
& \multicolumn{2}{c}{Edit Region} & \multicolumn{2}{c}{Background} & 3D Structure \\
\cmidrule(lr){2-3} \cmidrule(lr){4-5} \cmidrule(lr){6-6}
Variant & PSNR$\uparrow$ & LPIPS$\downarrow$ & PSNR$\uparrow$ & LPIPS$\downarrow$ & CD$\downarrow$ \\
\midrule
\textbf{JointEdit3D (full)} & \bestnum{26.82} & \bestnum{0.209} & \secondnum{30.69} & \secondnum{0.103} & \bestnum{0.1053} \\
\midrule
w/o main-state feedback to SceneAnchor & 24.35 & 0.312 & 29.73 & 0.114 & 0.1325 \\
Direct source conditioning (rank-16 LoRA) & 23.53 & 0.292 & \bestnum{30.88} & \bestnum{0.086} & 0.1194 \\
w/o source observation (zero SceneAnchor input) & 19.40 & 0.510 & 18.91 & 0.562 & 0.1829 \\
w/o GEO in Wan Flow (+VGGT) & 24.75 & 0.381 & 27.85 & 0.337 & 0.1506 \\
\midrule
w/o Text Condition & \secondnum{26.37} & \secondnum{0.214} & 30.66 & 0.109 & \secondnum{0.1064} \\
w/o Region Decomposition & 23.69 & 0.324 & 30.16 & 0.159 & 0.1257 \\
w/o Within-region Weighting & 25.35 & 0.234 & 29.94 & 0.119 & 0.1126 \\
w/o Temporal Weighting & 26.12 & 0.238 & 29.91 & 0.124 & 0.1148 \\
\bottomrule
\end{tabular}%
}
\endgroup
\end{table}

In Table~\ref{tab:ablation}, removing main-to-anchor feedback degrades editing and geometry, supporting SceneAnchor conditioning on the evolving denoising state. Replacing SceneAnchor with direct source-conditioning LoRA improves background preservation but weakens editing; its parameterization also differs. Without source video, background degrades most, suggesting the reference alone insufficiently constrains unedited content across views. Replacing joint geometry generation with RGB-only generation followed by VGGT degrades RGB and reconstruction but also changes the pretrained latent layout (Appendix~\ref{app:implementation_details}). Separating edit and background losses contributes most among loss variants, consistent with counteracting background dominance. Within-region and temporal weighting further help by emphasizing larger latent changes and positions farther from the reference in time. Text removal has less impact than removing source observations or region decomposition. Adding estimated source geometry worsens edit LPIPS and CD without helping background (Table~\ref{tab:rgb_geo_anchor}), so richer inputs need not improve conditioning. Source-corruption robustness is evaluated in Appendix~\ref{app:source_corruption}.

\section{Conclusion}
\label{sec:conclusion}

We presented JointEdit3D, a feed-forward framework that jointly predicts edited appearance and geometry in a pretrained unified RGB-geometry latent space. Asymmetric latent inpainting propagates a single reference edit along the source trajectory, while the SceneAnchor Branch and edit-aware losses balance edit fidelity with preservation of unedited content. We also introduced SceneEdit3D-15K and SceneEdit3D-Bench to support paired training and standardized evaluation of edit fidelity, background preservation, and 3D structure. Experiments show improved edited-region quality and structural completeness with competitive background preservation and efficient inference.

\clearpage

\subsection*{Reproducibility statement}
The model architecture and training objective are described in
Section~\ref{sec:method}. Dataset generation and annotations are detailed in
Section~\ref{sec:data_generation} and Appendix~\ref{app:dataset_details}, while
the training configuration, evaluation protocol, and metrics are reported in
Section~\ref{sec:exp_setup} and Appendices~\ref{app:implementation_details}, \ref{app:baseline_protocol}, and~\ref{app:3d_metric_protocol}.

\bibliography{references}
\bibliographystyle{iclr2027_conference}

\clearpage
\appendix
\setlength{\abovedisplayskip}{4pt plus 1pt minus 2pt}
\setlength{\belowdisplayskip}{4pt plus 1pt minus 2pt}
\setlength{\abovedisplayshortskip}{2pt plus 1pt}
\setlength{\belowdisplayshortskip}{2pt plus 1pt}
\newcommand{\appendixdatasetdetails}{%
\section{SceneEdit3D-15K Generation Details}
\label{app:dataset_details}

\subsection{Scene Sources and Candidate Extraction}
We build SceneEdit3D-15K from composable indoor scenes in Imaginarium~\citep{imaginarium}. Each scene is provided as a Blender file with object-level meshes, parent-child relations, physically based materials, lighting, camera metadata, and captions, so edits change the underlying 3D scene state rather than rendered images. We complete missing structural elements (e.g., walls and ceilings) to maintain complete indoor structure along the rendered camera trajectories, and we build an object table from scene metadata and rendered visibility statistics, recording each candidate's exact Blender name, semantic class, caption, parent, number of children, visible image area, maximum 3D extent, height, and support-likeness; structural objects (floors, ceilings, cameras, walls) are excluded from the editable set.

\subsection{VLM-Guided Edit Proposal}
We use GPT-5.5 as the VLM to propose canonical deletion and relocation tasks. Given scene renderings and a JSON layout file derived from the scene graph, the VLM analyzes object visibility, physical support, free space, and scene relations, then selects deletion and relocation targets with exact object names and proposes a camera-trajectory template. For relocation, Blender samples and validates the final target position under geometric constraints. Addition and move-reverse samples are obtained by reversing valid deletion and move pairs, and appearance changes are generated separately in Blender by replacing a target object's material or texture with a compatible alternative. Figure~\ref{fig:dataset_task_prompt} shows the task-proposal prompt; Figure~\ref{fig:dataset_refine_prompt} shows instruction refinement performed after rendering. These representative templates omit implementation-specific examples and repair logs for readability.

\begin{figure}[!tb]
\centering
\setlength{\fboxsep}{6pt}
\fcolorbox{black}{gray!6}{%
\begin{minipage}{0.92\linewidth}
\footnotesize
\textcolor{red!70!black}{\textbf{Representative task-proposal prompt}}\\[0.25em]
\textbf{System.} You are a senior 3D scene edit planner. Your output is consumed by an automatic Blender renderer, so object names, operation names, and JSON syntax must be exact. Return strict JSON only; do not output markdown or natural language outside JSON.\\[0.35em]
\textbf{Inputs.} Inspect the rendered scene views and the accompanying JSON layout file before selecting tasks. Use the renderings to judge visibility, occlusion, room boundaries, and visual plausibility, and use the JSON layout to reason about object names, support, containment, proximity, and parent-child relations.\\[0.35em]
\textbf{Task goal.} Generate exactly $K$ realistic, visually meaningful, and physically plausible canonical tasks. The allowed operations in this prompt are \texttt{delete} and \texttt{move}; do not generate \texttt{add}, \texttt{move\_reverse}, or appearance-change tasks here. These inverse or appearance tasks are produced later by the deterministic rendering/augmentation pipeline. Use only object names from the JSON layout file, copied character-for-character. Diversify object classes, scene regions, and object scales.\\[0.35em]
\textbf{Operation-specific rules.}
\textit{Delete:} choose a clearly visible and semantically meaningful removable object; avoid structural elements, room-defining objects that are too large, tiny objects whose removal is visually ambiguous, and support objects with many children unless deleting the whole group is plausible. Prefer objects whose removal creates a visible but physically plausible empty region.
\textit{Move:} choose a non-support leaf object with visible free space nearby; move it only a moderate distance within the same functional area, keep it supported after moving, avoid collisions or wall penetration, and ensure the object is visible both before and after the move.
\textit{Appearance:} do not select appearance tasks in this prompt; material/texture changes are sampled by Blender from compatible material or texture candidates.\\[0.35em]
\textbf{Trajectory rule.} Select one template from \texttt{trajectory\_templates}. The selected trajectory should view the affected objects clearly and avoid severe occlusion; include a short reason.\\[0.25em]
\textbf{Output schema.}\\[-0.25em]
{\ttfamily\footnotesize
\{``tasks'': [\{``operation'': ``delete|move'',\\
\hspace*{1.1em}``target\_object'': ``exact\_object\_name'',\\
\hspace*{1.1em}``edit\_text\_en'': ``short edit description'',\\
\hspace*{1.1em}``instruction'': ``renderer-facing concise instruction'',\\
\hspace*{1.1em}``trajectory\_reason'': ``why this view works'',\\
\hspace*{1.1em}``trajectory\_template'': ``one provided template''\}]\}
}
\end{minipage}}
\caption{Representative VLM prompt for proposing canonical renderer-executable tasks. The VLM analyzes rendered scene views and a JSON layout file before selecting deletion and relocation targets under physical and visibility constraints. Add and move-reverse samples are derived later by state/time reversal; appearance changes use Blender-side material/texture replacement.}
\label{fig:dataset_task_prompt}
\end{figure}

\begin{figure}[!tb]
\centering
\setlength{\fboxsep}{6pt}
\fcolorbox{black}{gray!6}{%
\begin{minipage}{0.92\linewidth}
\footnotesize
\textcolor{red!70!black}{\textbf{Editing-instruction refinement prompt}}\\[0.25em]
\textbf{Role.} You are generating natural editing instructions to fine-tune a video diffusion model for scene editing. You are shown three rendered inputs from the same viewpoint: \textbf{before editing}, \textbf{after editing}, and an \textbf{edit-region mask} where white marks the changed area. You also receive the operation type and a draft renderer instruction.\\[0.35em]
\textbf{Visual analysis.} Compare the before/after images and use the mask only to localize the change. Identify the edited object, its appearance, nearby landmarks, support surface, and the visible result after the edit. Do not mention the mask, frame index, image pair, renderer, or any technical term in the final instruction.\\[0.35em]
\textbf{Prompt style.} Generate one forward instruction and one reverse instruction. Each prompt should be 15--40 words, start with an action verb such as remove, add, move, or replace, include spatial context, describe the visual outcome, and include color/material/size cues when visible.\\[0.25em]
\textbf{Output schema.}\\[-0.2em]
{\ttfamily\footnotesize
\{``forward'': ``before-to-after editing prompt'',\\
\hspace*{1.1em}``reverse'': ``after-to-before editing prompt''\}
}
\end{minipage}}
\caption{Representative prompt for converting rendered before/after evidence into natural user-facing editing instructions. This stage grounds the language in the actual rendered change rather than only in object identifiers.}
\label{fig:dataset_refine_prompt}
\end{figure}

Raw VLM responses are validated rather than accepted directly: each parsed task is checked against the candidate table and layout and is rejected for unknown object names, repeated targets, violated operation-specific constraints, deletion targets that are too small or structurally important, or move targets without sufficient nearby free space and support. When too few tasks are valid, a repair prompt returns the rejection reasons and the previous response, asking the VLM to keep valid tasks and replace only invalid ones; remaining gaps are filled by rule-based diversification over the validated candidate pool.

\subsection{Deterministic Scene Execution}
Accepted canonical tasks are executed in Blender under a deterministic before/after protocol, and inverse tasks are derived by state/time reversal. Deletion hides the target object and its descendants in the edited scene (child lights are preserved to keep light-source settings fixed); the addition sample swaps the source and edited states and reverses the rendered sequence. Relocation keeps the target visible in both scenes and moves it to a new collision-free position within the navigable room bounds, with the reverse sample derived by swapping states and reversing frame order. Appearance changes are sampled by Blender without VLM proposal, replacing a material or texture with a compatible alternative while preserving mesh and transform. Multi-op samples apply two or more atomic operators to \emph{distinct} objects in one execution; same-object conflicts (e.g., move then delete the same instance) are rejected, and the camera path is centered on the affected set.

\begin{figure}[!tb]
\centering
\setlength{\fboxsep}{6pt}
\fcolorbox{black}{gray!6}{%
\begin{minipage}{0.92\linewidth}
\small
\textbf{Algorithm 1: Deterministic Blender Execution for One Edit}\\[0.25em]
\textbf{Input:} Blender scene $\mathcal{S}_0$, metadata $\mathcal{M}$, edit task $q$, JSON layout file $\mathcal{L}$, seed $s$. For an atomic edit, $q$ is a validated delete/move proposal or a Blender-sampled appearance task.\\
\textbf{Output:} paired source/target videos, masks, depth maps, camera poses, intrinsics, and task metadata.\\[-0.2em]
\begin{enumerate}
\setlength{\itemsep}{0.15em}
\item Parse the JSON layout file $\mathcal{L}$ and scene metadata $\mathcal{M}$ to recover object identities, support relations, parent-child links, approximate object extents, and navigable room bounds.
\item Resolve affected object roots from $q$, expand them with the parent-child graph, and run layout validity checks to reject structural, too-small, or unsupported targets.
\item For relocation, sample candidate target positions and run collision detection, support-surface checks, room-bound checks, and free-space validation before accepting a placement.
\item Generate a seeded camera trajectory $P=\{(R_i,t_i)\}_{i=1}^{N}$ from fixed templates such as horizontal sweeps, diagonal motions, rising motions, and drop-forward motions, centered on the affected object set.
\item Run operation-specific visibility checks along $P$: deletion requires target visibility before editing, while relocation and appearance changes require visibility in both states. Reject or repair tasks that fail these checks; inverse additions inherit visibility from the deletion source.
\item Save the initial object states and camera state so both branches start from the same scene.
\item For each branch $b \in \{\mathrm{before},\mathrm{after}\}$: restore the saved scene state, apply the branch-specific edit operator $E_q^b$, attach the same camera path $P$, and render RGB/depth frames.
\item Select branch-specific mask objects: deleted objects for deletion, moved objects in both relocation branches, and appearance-changed objects in both appearance branches. Render binary masks through Blender object-index passes.
\item Export RGB videos, depth maps, masks, relative camera poses, intrinsics, task JSON, and completion sentinels.
\item Derive inverse samples for deletion and relocation by swapping before/after branches and reversing frame, depth, mask, and pose order: delete $\rightarrow$ add and move $\rightarrow$ move-reverse.
\item Discard samples with missing files, corrupted frames, invalid physical placement, failed operation-specific visibility checks, or empty masks in branches where the edited object should be visible.
\end{enumerate}
\vspace{-0.2em}
\[
E_q^b(\mathcal{S}_0)=
\begin{cases}
\mathrm{hide}(o), & q=\mathrm{delete},\, b=\mathrm{after},\\
\mathrm{move}(o,\Delta), & q=\mathrm{move},\, b=\mathrm{after},\\
\mathrm{replace\_mat}(o,m'), & q=\mathrm{appearance},\, b=\mathrm{after},\\
\mathrm{identity}, & \text{otherwise}.
\end{cases}
\]
\end{minipage}}
\caption{Algorithmic view of the deterministic Blender execution protocol. The operators shown define atomic edits; multi-operation tasks compose operators on distinct objects. Both rendering branches reuse the same accepted camera path and render settings to produce paired observations under identical viewpoints.}
\label{fig:blender_execution_algorithm}
\end{figure}

Camera trajectories are generated from fixed templates (horizontal sweeps, diagonal forward, rising, and drop-forward motions) centered on the edited object or, for composite edits, on the midpoint of the affected objects. Relocation candidates are tested for room-bound validity, collision, support plausibility, and visibility from the final path, and tasks whose targets are poorly visible are rejected or repaired, so the source and target videos differ primarily by the intended edit rather than by accidental occlusion or camera drift.

\subsection{Rendering, Annotations, and Prompt Refinement}
Every accepted edit is rendered as a source video and an edited target video with identical camera poses, intrinsics, frame count, resolution, lighting configuration, and sampling settings, so frame $i$ of the two sequences forms a strictly paired before/after observation. We additionally export camera intrinsics, relative poses, depth maps, and task metadata. The edited reference frame $I_e$ is selected from the target sequence and paired with the final instruction $p$ as the conditioning input used by JointEdit3D.

Edit masks are rendered through Blender object-index passes and track the affected object: deleted or relocated objects, the added object in the edited branch, or the material-replaced object; for composite edits, branch-specific masks are combined when a single edit-region mask is needed for prompt refinement or evaluation. Because the masks come from the renderer, they align exactly with the RGB and depth frames and require no separate segmentation model.

The final language instruction is refined after rendering: a VLM receives the edited reference frame, its source counterpart at the same viewpoint, and the combined mask, and generates forward and reverse prompts in a compact JSON format. The refined instructions describe the rendered edit and its visible outcome, converting renderer-facing identifiers into natural language with spatial context and appearance cues. For example, ``remove object chair\_12'' becomes ``Remove the brown lounge chair beside the sofa, revealing the wooden floor and wall behind it.'' The before/after pair and mask support offline instruction generation; the editor receives the source RGB video, edited reference, and final instruction.

\subsection{Dataset Organization and Release Format}
\label{app:dataset_release_schema}

The dataset is organized into separate JSONL manifests for training, validation, and test. Each entry records a sample ID, RGB/mask and geometry shard keys, the edited-reference frame index, an edit prompt, and the number of frames. RGB frames and masks are stored in one shard family; depth maps and camera parameters are stored in the corresponding geometry shards under the same sample ID. The dataset license is CC BY-NC-SA 4.0 with attribution to the Imaginarium~\citep{imaginarium} authors, matching the source scenes' ShareAlike terms.

\subsection{Training and Evaluation Data Flow}
SceneEdit3D-15K separates the RGB pairs used for JointEdit3D training from the renderer-side 3D annotations released with the dataset. Each paired source video $\mathcal{V}_{\mathrm{src}}$ and edited target video $\mathcal{V}_{\mathrm{edit}}$ is converted into the unified RGB-geometry latent space of Gen3R~\citep{gen3r}: the RGB half comes from the Wan video VAE, $\mathbf{z}_{\mathrm{edit}}^{\mathrm{rgb}}=E_{\mathrm{rgb}}(\mathcal{V}_{\mathrm{edit}})$, and the geometry half from the frozen VGGT encoder $\Phi_{\mathrm{vggt}}$~\citep{vggt} and Geometry Adapter encoder, $\mathbf{z}_{\mathrm{edit}}^{\mathrm{geo}}=A_{\psi}^{\mathrm{enc}}(\Phi_{\mathrm{vggt}}(\mathcal{V}_{\mathrm{edit}}))$. Because VGGT predicts camera-aware multi-view geometry features from RGB and the Geometry Adapter maps them onto the Wan latent grid, this yields a geometry-latent pseudo-target aligned with the RGB latent in time and space, giving $\mathbf{z}_{\mathrm{edit}}=[\mathbf{z}_{\mathrm{edit}}^{\mathrm{rgb}};\mathbf{z}_{\mathrm{edit}}^{\mathrm{geo}}]$. The source video is encoded in the same space; its RGB half feeds the SceneAnchor Branch, and its full latent is used on the loss side to compute source-target latent-difference masks and weights.

During training, the model is conditioned on the source RGB video, the edited reference frame, and the language instruction, while $\mathbf{z}_{\mathrm{edit}}$ provides the flow-matching supervision: we corrupt it with Gaussian noise and train the editor to predict the flow-matching velocity for both its RGB and VGGT-derived geometry halves. Gradients update only the SceneAnchor Branch; the Wan VAE, VGGT encoder, Geometry Adapter, and Gen3R-tuned backbone remain frozen. The frozen VGGT--Geometry-Adapter encoding pipeline converts the edited RGB target video into geometry-latent supervision. JointEdit3D thus derives appearance and geometry-latent supervision from paired RGB videos while preserving the pretrained joint representation, without using renderer-provided depth maps, point clouds, or camera poses as direct training targets.

Renderer annotations support data preparation, calibration, and evaluation. Object masks are used for visibility and edit-area filtering, instruction refinement, reference-frame selection, and calibration of the latent-difference threshold $\tau$. At evaluation, they define the object and background regions for image metrics. Depth maps, camera intrinsics/extrinsics, and edited scene states provide geometric references for depth and point-cloud comparisons. The training objective uses RGB latents and VGGT-derived geometry latents as targets, with region weights computed from source--target latent differences.

\subsection{Quality Control and Benchmark Curation}
Quality control runs at three stages: candidate objects are filtered by semantic validity, visibility, size, parent-child structure, and operation-specific safety before rendering; output completeness is checked, with completion sentinels written only after RGB frames, videos, masks, depth maps, poses, intrinsics, and metadata are generated; and samples with missing frames, corrupted files, failed operation-specific visibility checks, or empty or overly small edit regions in branches where the affected object should be visible are removed afterwards.

As Table~\ref{tab:bench_composition} reports, SceneEdit3D-15K contains 15,319 paired samples: 13,799 training samples from 135 scenes and 1,520 held-out samples from 16 scenes absent from training. The held-out pool is partitioned into 1,420 validation samples and 100 test samples. Training and held-out data are scene-disjoint; validation and test are sample-disjoint subsets of the held-out pool, with no forward/reverse counterparts of the same edit shared across the two splits.

The test split, SceneEdit3D-Bench, draws from 15 held-out scenes and covers all five operation groups, with more samples allocated to addition and deletion. Edited-object area is measured as the renderer-defined edit-mask area divided by the image area in a randomly selected frame specified in each sample's dataset annotations. The benchmark spans sparse (${<}1\%$) to large (${\geq}15\%$) edited regions and includes partially occluded, small-object, large-object, and multi-operation cases, supporting evaluation of edit fidelity, background preservation, and 3D structure under a common paired-rendering protocol.

\begin{table}[!tb]
\centering
\caption{SceneEdit3D-15K split statistics and SceneEdit3D-Bench composition. Top: sample, scene, and operation counts; Test is a subset of the Held-out pool. Bottom: edit-mask image-area ratios measured in the randomly selected frame specified in each sample's dataset annotations.}
\label{tab:bench_composition}
\begingroup
\small
\setlength{\tabcolsep}{4.5pt}
\renewcommand{\arraystretch}{1.08}
\begin{tabular}{@{}l c c c c c c c@{}}
\toprule
Split & Total & Scenes & Add & Delete & Move & Appearance & Multi-op \\
\midrule
Train & 13,799 & 135 & 4,990 & 4,987 & 1,884 & 1,017 & 921 \\
Held-out pool & 1,520 & 16 & 600 & 600 & 186 & 24 & 110 \\
Test (Bench) & 100 & 15 & 29 & 29 & 14 & 14 & 14 \\
\bottomrule
\end{tabular}

\vspace{0.5em}

\begin{tabular}{@{}l cccc@{}}
\toprule
Edited-object area ratio & $<1\%$ & $1$--$5\%$ & $5$--$15\%$ & $\geq15\%$ \\
\midrule
Count & 11 & 47 & 29 & 13 \\
\bottomrule
\end{tabular}
\endgroup
\end{table}

\FloatBarrier
}

\section{Response Diagnostics and Calibration}
\label{app:loss_diagnostic}

This section visualizes the complementary latent cues used for region-decomposed supervision, the denoiser response to the edited reference, and threshold calibration. The latent-difference masks are used only during training.

\begin{nolinenumbers}
\begin{center}
\includegraphics[width=0.88\textwidth]{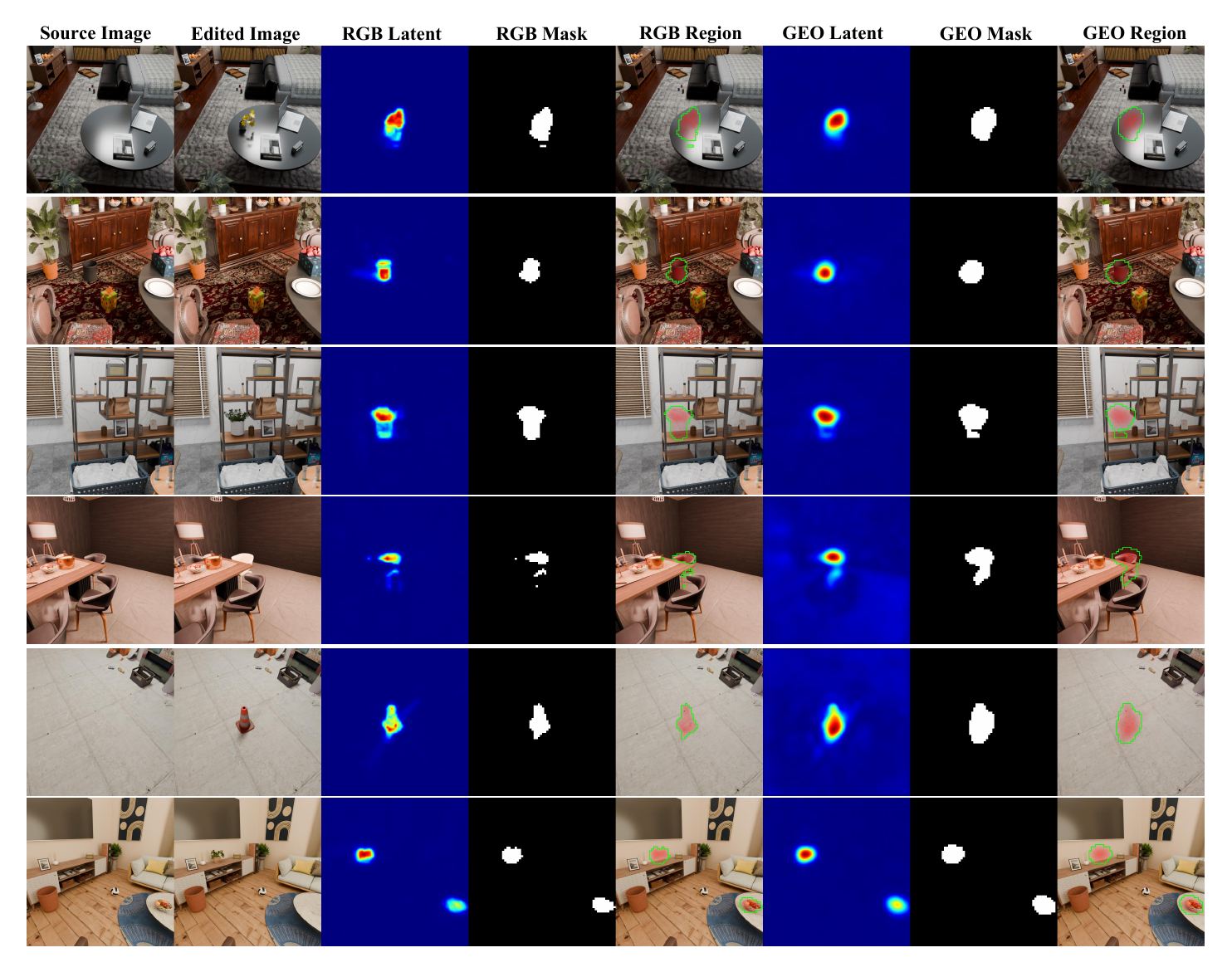}
\captionof{figure}{Latent-difference visualization for region-decomposed supervision. RGB and geometry latent deltas provide complementary edit cues, which are converted into latent masks for separating edited and background regions.}
\label{fig:training_loss_diagnostic}
\end{center}
\end{nolinenumbers}

Geometry deltas emphasize structural changes around the edited object, while RGB deltas capture appearance changes that extend beyond its silhouette. These complementary cues allow supervision to account for both the object edit and its surrounding effects. We therefore derive training masks from latent differences and assign larger within-region weights to positions with larger changes.

\subsection{Edit-Condition Impact Visualization}
\label{app:edit_condition_impact}

\begin{figure}[!tb]
\centering
\includegraphics[width=\textwidth]{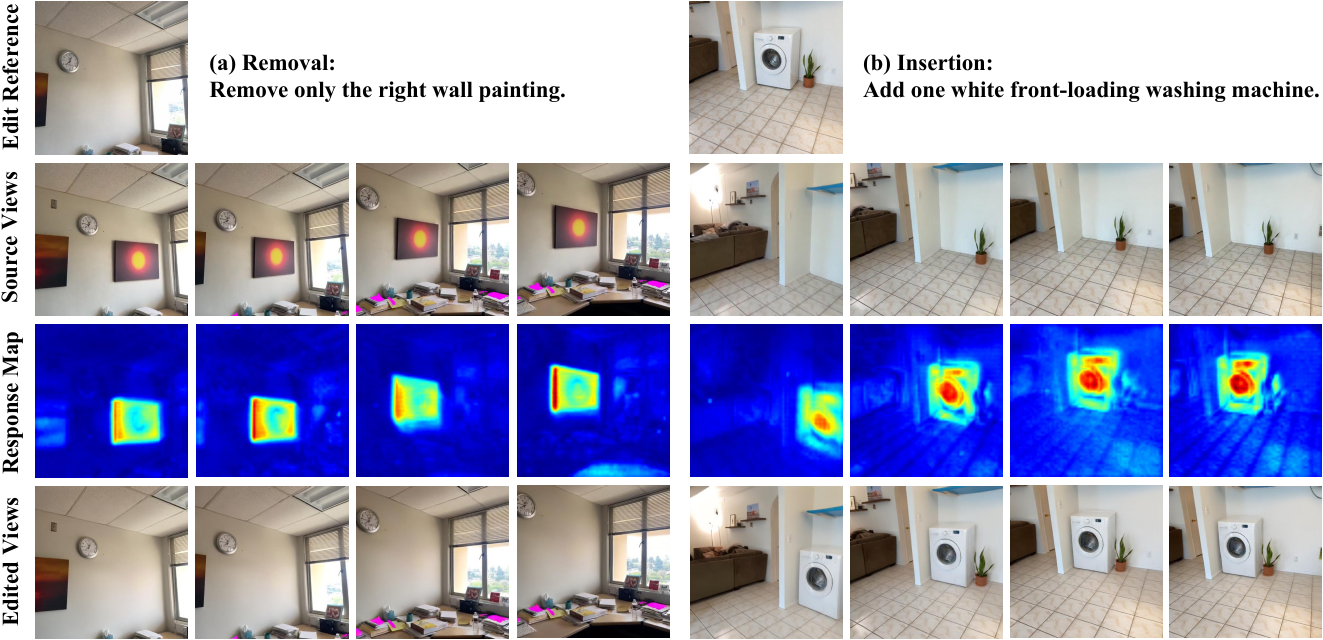}
\caption{Edit-condition impact visualization. Heatmaps show the denoiser response difference with and without the edited reference condition, keeping the prompt and source-video branch fixed. Warmer regions indicate where the edited reference condition most strongly changes the model prediction.}
\label{fig:inference_response}
\end{figure}

To visualize where the edited reference affects generation, we fix the same noisy latent, timestep, prompt, and source-video control branch, and compare two denoising forward passes: one with the full edit condition and one with the edited reference condition removed. We compute
\[
\Delta_{\mathrm{edit}}
= G_{\theta,\pi}(z_t, c_{\mathrm{source}}, c_{\mathrm{edit}})
- G_{\theta,\pi}(z_t, c_{\mathrm{source}}, c_{\mathrm{no\,ref}}),
\qquad
M_{\mathrm{edit}} = \mathrm{mean}_c(|\Delta_{\mathrm{edit}}|),
\]
then upsample and normalize \(M_{\mathrm{edit}}\) to obtain the heatmap shown in Figure~\ref{fig:inference_response}.

Both passes retain the source-video branch and prompt; $c_{\mathrm{no\,ref}}$ denotes removal of the edited reference condition only. The heatmaps show that the edited reference guides denoiser responses toward the intended edit and nearby affected context.

\subsection{Threshold Calibration for Edit/Background Separation}
\label{app:threshold_calibration}

We calibrate the binary edit-region threshold on a held-out subset with renderer-provided object masks. Let \(G\) denote the affected-object mask. For each threshold \(\tau\), we obtain a predicted edit mask from the normalized latent difference and measure object-region coverage and responses outside the object mask:
\[
\begin{alignedat}{2}
P_\tau &= \mathbf{1}[\mathrm{diff\_norm} > \tau], \qquad&
\mathrm{Coverage} &= \frac{|P_\tau \cap G|}{|G|},\\
\mathrm{Overflow} &= \frac{|P_\tau \setminus G|}{|G|}, \qquad&
\mathrm{Score} &= \frac{\mathrm{Coverage}}{1+\mathrm{Overflow}}.
\end{alignedat}
\]
For each mask pair, Score equals the intersection-over-union \(|P_\tau \cap G|/|P_\tau \cup G|\). Overflow measures responses outside the object mask, including valid secondary effects; it is a spatial calibration measure rather than an editing-error rate. We set \(\tau=0.15\), near the coverage--overflow elbow, retaining high object coverage while limiting spatial spillover. Larger thresholds suppress weaker edit responses.

\begin{figure}[!tb]
\centering
\includegraphics[width=0.72\textwidth]{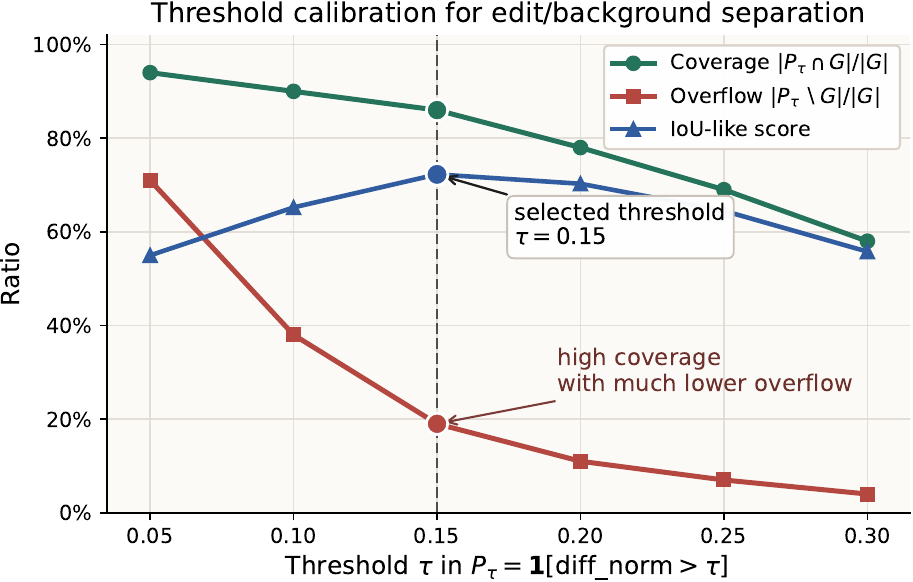}
\caption{Threshold calibration for latent-difference edit/background decomposition over \(\tau \in \{0.05,0.10,0.15,0.20,0.25,0.30\}\). We choose \(\tau=0.15\), near the coverage--overflow elbow.}
\label{fig:coverage_overflow_elbow}
\end{figure}

\FloatBarrier

\section{Additional Experimental Results}

\subsection{Implementation Details}
\label{app:implementation_details}

\paragraph{Input representation and conditioning.}
All videos are represented as 49 RGB frames at \(560{\times}560\) resolution. The Wan video VAE encodes each sequence into 16-channel RGB latents with 13 temporal steps and \(70{\times}70\) spatial size. Following Gen3R, the geometry stream is encoded by frozen VGGT and Geometry Adapter modules; we concatenate RGB and geometry latents along the width dimension, producing a target latent of shape \(16{\times}13{\times}70{\times}140\). Each sample provides one edited reference frame \(I_e\) at zero-based clip index \(r\in\{0,\ldots,48\}\). For synthetic data, this frame is selected from views with high edit-mask coverage. Under Wan's causal temporal packing, latent step 0 corresponds to the first frame, while step \(j\geq1\) packs frames \(4j-3\) through \(4j\). Thus, \(k=\lceil r/4\rceil\), including \(k=0\) for \(r=0\). We encode \(I_e\) alone and place its latent at step \(k\); the clip is not reordered. To construct the four-channel observation mask, we mark frame \(r\) in pixel time, repeat the first-frame entry four times, and pack consecutive groups of four into channels. For \(r=0\), all four channels at step 0 are marked; for \(r>0\), only channel \((r-1)\bmod4\) at step \(k\) is marked. These channels encode a single observed RGB frame. Remaining RGB steps and the full geometry half stay masked. The source-video RGB latent, concatenated with a zero geometry half, serves as the SceneAnchor Branch control input.

The single-image encoding is an auxiliary condition in $\mathbf{y}$ (Equation~\ref{eq:y_concat}); it is not substituted into the noisy state $\mathbf{z}_t$ or the target video latent. Training supervises the full edited-video latent.

\paragraph{Training.}
For the full JointEdit3D model, we train only the SceneAnchor Branch and keep the Gen3R-tuned Wan transformer, VAE, text encoder, CLIP image encoder, VGGT, and Geometry Adapter frozen. Cached latents and text/image features are used to reduce training cost. The main model is trained for 10K steps on 8 NVIDIA H200 GPUs with batch size 10 per device, bf16 mixed precision, gradient checkpointing, AdamW with learning rate \(10^{-4}\), cosine decay, and EMA. We apply light condition dropout to the edited image and text prompt to improve robustness.

The main results use the 10K-step model. Due to computational constraints, all component ablations in Table~\ref{tab:ablation} use a common 2K-step schedule, including the full-model baseline.

\paragraph{Architecture and ablations.}
SceneAnchor blocks are injected at layers \(\mathcal{S}=\{0,4,8,12,16,20,24,28\}\). The ``Direct source conditioning (rank-16 LoRA)'' variant in Table~\ref{tab:ablation} replaces SceneAnchor with direct source conditioning and adapts the frozen Gen3R-tuned transformer using rank-16 LoRA. Both designs are trained for 2K steps, with different conditioning pathways and trainable parameterizations: 70M SceneAnchor parameters for the full model versus LoRA adapters for direct conditioning. Appendix~\ref{app:main_state_feedback} separately tests main-state feedback within the SceneAnchor architecture. The ``w/o GEO in Wan Flow (+VGGT)'' variant generates only RGB latents, reducing the grid from $16\times13\times70\times140$ to $16\times13\times70\times70$, and reconstructs its RGB output using the VGGT backend from Table~\ref{tab:external_reconstruction}. This ablation changes both the generated modalities and the pretrained latent layout.

\paragraph{Temporal weighting.}
The training-only temporal weight is
\begin{equation}
\label{eq:temporal_weight}
a_i = 1 + \eta\frac{|\rho(i)-k|}{\max_j |\rho(j)-k| + \varepsilon_{\mathrm{den}}},
\qquad
\tilde{l}_i = a_i l_i ,
\end{equation}
where $\rho(i)$ is the latent temporal index of position $i$ and $k$ is the reference step. The weight emphasizes positions farther from the reference in time, with \(a_i\in[1,1.5]\) for \(\eta=0.5\). Temporal distance is a proxy for viewpoint change and does not account for revisits or loop closures. The remaining loss settings are \(\tau{=}0.15\), \(\alpha{=}5.0\), \(\lambda{=}0.5\), and \(\gamma_{\mathrm{geo}}{=}1.5\).

\paragraph{Inference.}
Inference uses the same 49-frame preprocessing and single-reference conditioning. Unless otherwise stated, quantitative evaluation uses FlowMatch Euler sampling with 5 denoising steps, classifier-free guidance scale 1.0, control scale 1.0, seed 42, and zero camera embeddings. Zero camera embeddings disable explicit camera guidance; the ordered source RGB observations provide trajectory evidence through SceneAnchor, alongside the reference and pretrained model priors. Camera predictions are outputs rather than input pose constraints. After denoising, the RGB half is decoded by the Wan VAE, and the geometry half is decoded by the Geometry Adapter to recover depth, point maps, valid masks, and camera predictions.

\subsection{Hyperparameter Sensitivity}
\label{app:hyperparameter_sensitivity}

\paragraph{Parameter and checkpoint selection.}
Loss hyperparameters and the checkpoint selection rule are fixed before evaluation, with $\tau$ calibrated separately in Appendix~\ref{app:threshold_calibration}. Table~\ref{tab:hyperparameter_sensitivity} varies one hyperparameter at a time while keeping the others at their defaults. All variants follow the 2K-step protocol of Table~\ref{tab:ablation} and are evaluated on the 100-sample SceneEdit3D-Bench test set.

\begin{table}[htbp]
\centering
\caption{Hyperparameter sensitivity on SceneEdit3D-Bench. Each block varies one objective hyperparameter around the default setting used by JointEdit3D.}
\label{tab:hyperparameter_sensitivity}
\begingroup
\small
\setlength{\tabcolsep}{4pt}
\renewcommand{\arraystretch}{1.04}
\begin{tabular}{@{}l c cc cc c@{}}
\toprule
& & \multicolumn{2}{c}{Edit Region} & \multicolumn{2}{c}{Background} & 3D Structure \\
\cmidrule(lr){3-4} \cmidrule(lr){5-6} \cmidrule(lr){7-7}
Hyperparameter & Value & PSNR$\uparrow$ & LPIPS$\downarrow$ & PSNR$\uparrow$ & LPIPS$\downarrow$ & CD$\downarrow$ \\
\midrule
\multirow{3}{*}{Within-region scale $\alpha$}
& 2.5 & 25.71 & 0.252 & \secondnum{30.02} & 0.129 & 0.1184 \\
& \textbf{5.0} & \bestnum{26.82} & \bestnum{0.209} & \bestnum{30.69} & \bestnum{0.103} & \bestnum{0.1053} \\
& 7.5 & \secondnum{26.18} & \secondnum{0.228} & 29.93 & \secondnum{0.126} & \secondnum{0.1136} \\
\midrule
\multirow{3}{*}{Temporal scale $\eta$}
& 0.25 & 25.96 & 0.239 & 29.87 & 0.131 & 0.1161 \\
& \textbf{0.50} & \bestnum{26.82} & \bestnum{0.209} & \bestnum{30.69} & \bestnum{0.103} & \bestnum{0.1053} \\
& 0.75 & \secondnum{26.20} & \secondnum{0.229} & \secondnum{30.05} & \secondnum{0.123} & \secondnum{0.1124} \\
\midrule
\multirow{3}{*}{Edit/background balance $\lambda$}
& 0.40 & 25.89 & 0.247 & \bestnum{30.91} & \bestnum{0.096} & 0.1145 \\
& \textbf{0.50} & \bestnum{26.82} & \bestnum{0.209} & \secondnum{30.69} & \secondnum{0.103} & \bestnum{0.1053} \\
& 0.60 & \secondnum{26.47} & \secondnum{0.221} & 29.86 & 0.132 & \secondnum{0.1118} \\
\midrule
\multirow{3}{*}{Geometry loss weight $\gamma_{\mathrm{geo}}$}
& 1.0 & \secondnum{26.19} & \secondnum{0.231} & \secondnum{30.12} & \secondnum{0.124} & 0.1208 \\
& \textbf{1.5} & \bestnum{26.82} & \bestnum{0.209} & \bestnum{30.69} & \bestnum{0.103} & \bestnum{0.1053} \\
& 2.0 & 25.98 & 0.238 & 29.95 & 0.128 & \secondnum{0.1109} \\
\bottomrule
\end{tabular}
\endgroup
\end{table}

The default settings achieve the best edited-region quality and geometry accuracy across the tested values. Reducing $\lambda$ favors background preservation at the expense of edit fidelity, reflecting its role in balancing the two regions. Intermediate values of $\alpha$, $\eta$, and $\gamma_{\mathrm{geo}}$ outperform both lower and higher settings, supporting balanced emphasis on latent changes, temporal distance, and geometry supervision.

\subsection{Additional Qualitative Comparisons}
\label{app:additional_qualitative}

Relocation, multi-operation, and deletion comparisons are in Figures~\ref{fig:qualitative} and~\ref{fig:qualitative_deletion}. The figures below add appearance, occluded removal, multi-reference editing, and native point clouds on DAVIS, ScanNet++, CO3D, and casual captures~\citep{davis, scannetpp, co3d}. Unless otherwise stated, real-scene examples use a cropped clip, one randomly selected keyframe, and a Nano Banana edit of that frame, demonstrating transfer from synthetic training to real captures.

\begin{figure}[!tb]
\centering
\includegraphics[width=\textwidth]{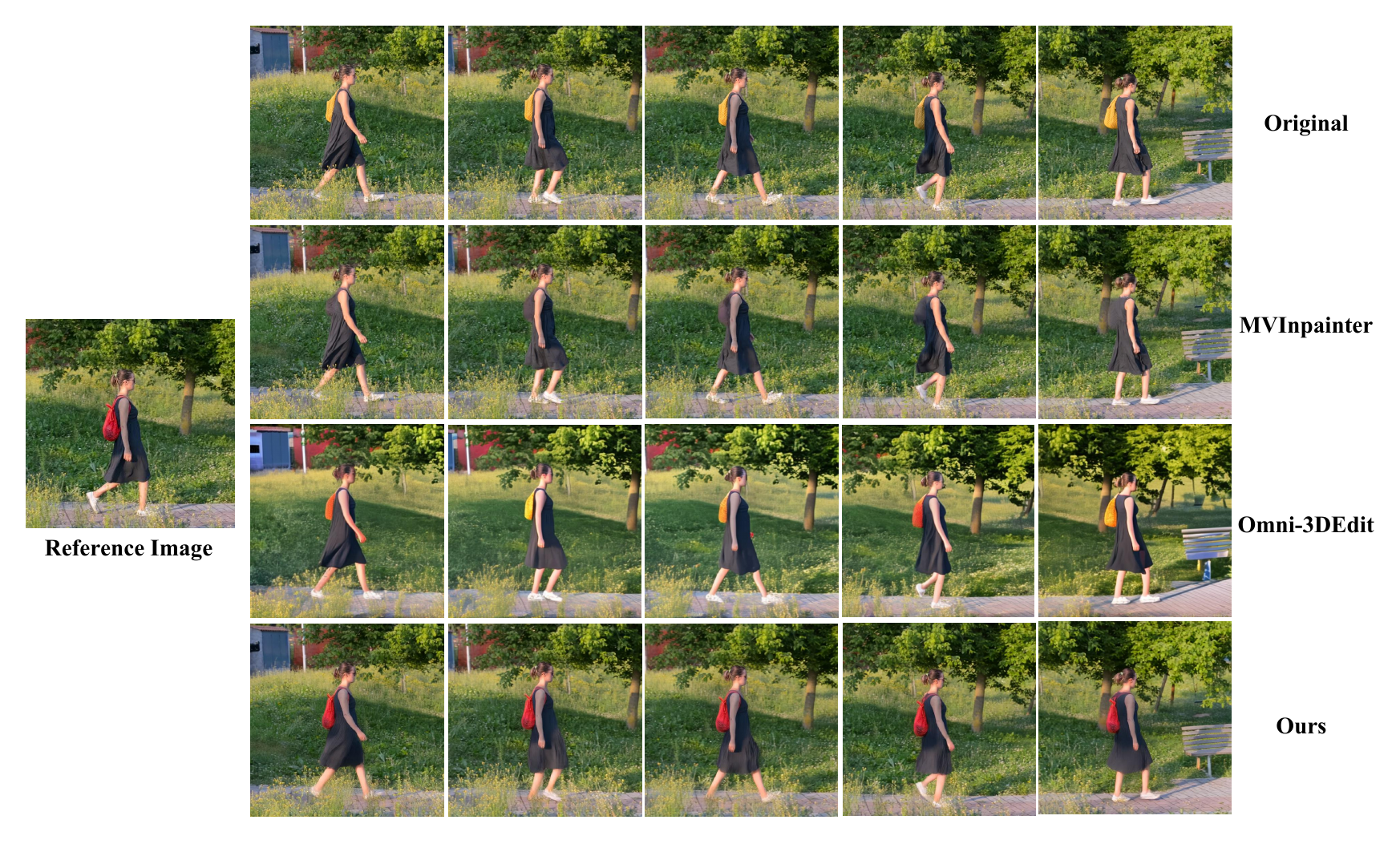}
\caption{Reference-guided backpack recoloring on DAVIS.}
\label{fig:appendix_appearance_edit_qualitative}
\end{figure}

\begin{figure}[!tb]
\centering
\includegraphics[width=1\textwidth]{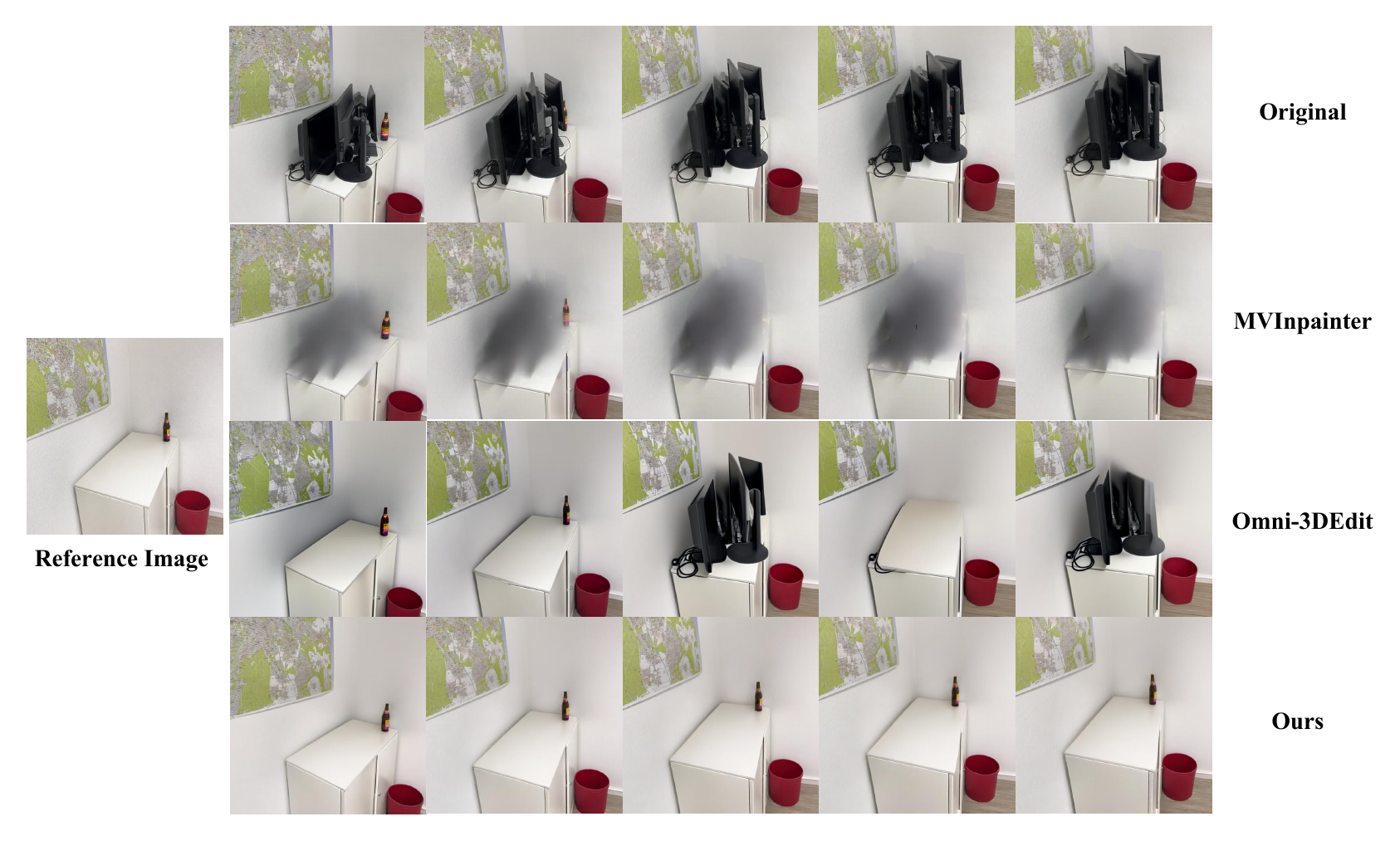}
\caption{Monitor removal on ScanNet++ with preservation of the bottle behind the monitors.}
\label{fig:appendix_real_scene_removal_qualitative}
\end{figure}

Figure~\ref{fig:appendix_appearance_edit_qualitative} recolors a backpack on DAVIS. All methods receive the same edited reference; MVInpainter additionally receives a SAM3 backpack mask, while Omni-3DEdit and JointEdit3D use no input mask. MVInpainter removes the backpack instead of recoloring it. Omni-3DEdit produces inconsistent backpack colors across frames and alters the background appearance. JointEdit3D maintains the red backpack across views while preserving the person and background.

Figure~\ref{fig:appendix_real_scene_removal_qualitative} removes a monitor cluster on ScanNet++ while keeping the bottle behind it. MVInpainter replaces the cluster with blur, while Omni-3DEdit leaves monitors or cables in some views. JointEdit3D removes the foreground monitors, preserves the bottle behind them, and restores continuous cabinet and wall surfaces across views.

\begin{figure}[!tb]
\centering
\includegraphics[width=1\textwidth]{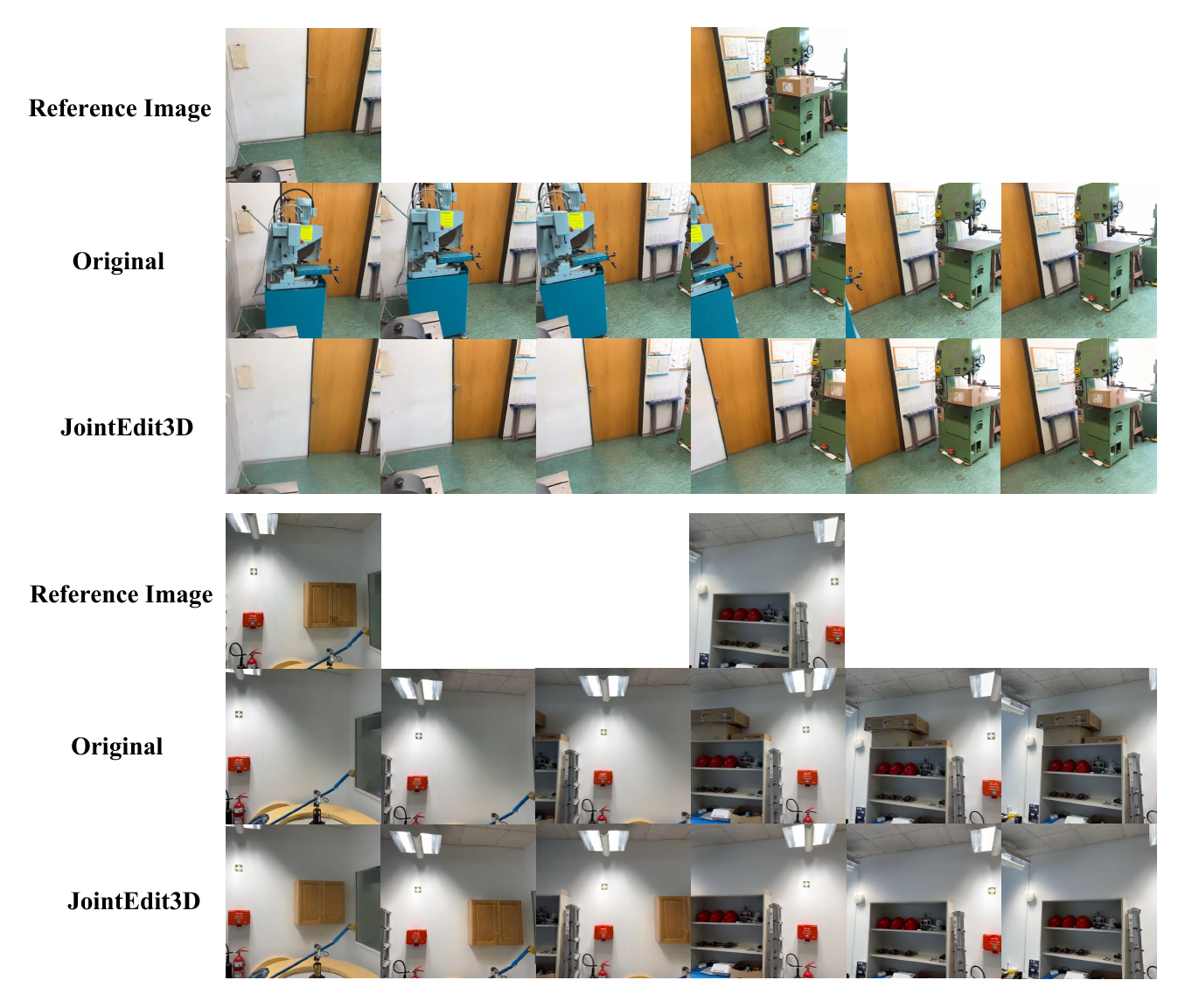}
\caption{Multi-reference editing in real scenes: removing a machine and adding a box (top), and adding a wall cabinet and removing boxes above the shelving unit (bottom).}
\label{fig:appendix_real_scene_multi_edit_qualitative}
\end{figure}

Although trained with a single edited reference, JointEdit3D accepts multiple edited references at inference by placing their encodings at the corresponding conditioning positions. Figure~\ref{fig:appendix_real_scene_multi_edit_qualitative} demonstrates combined removal and insertion across different parts of a room, propagating both reference edits while preserving the surrounding scene.

\begin{figure}[!tb]
\includegraphics[width=1\textwidth]{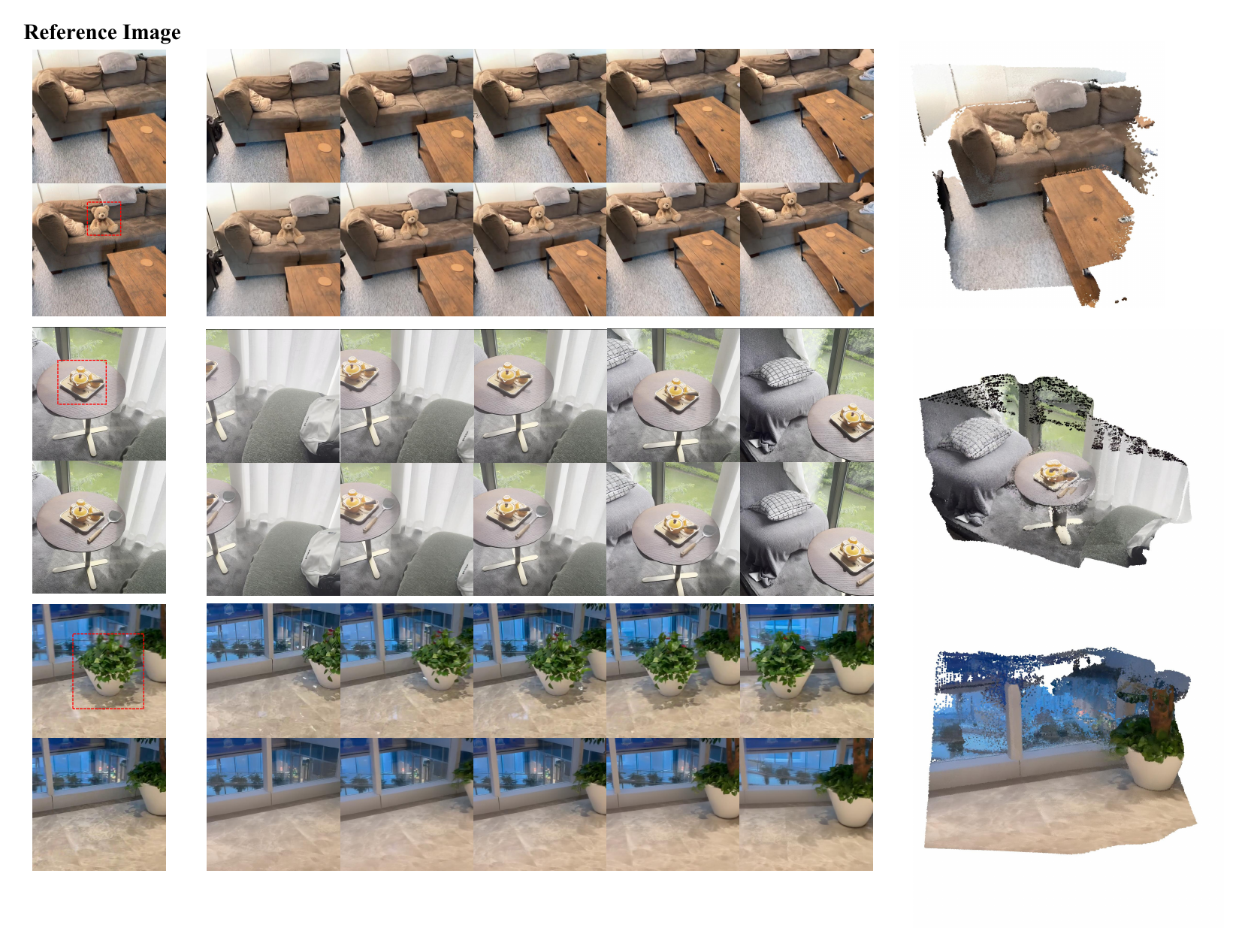}
\caption{Real-scene editing with jointly generated RGB views and native point clouds on ScanNet++ and casual indoor captures.}
\label{fig:appendix_real_scene_desk_qualitative}
\end{figure}

Figure~\ref{fig:appendix_real_scene_desk_qualitative} shows the source clip, the edited sequence, and the decoded point cloud on ScanNet++ and casual indoor captures. Object additions and removals appear consistently in the generated views and native point clouds, while the surrounding furniture layout is preserved.

\subsection{Operation-Specific Diagnostics}
\label{app:operation_specific}

Table~\ref{tab:operation_specific} complements the main evaluation with operation-wise SSIM and depth metrics. JointEdit3D achieves depth threshold accuracy above 93\% across all five operations. Relocation has the largest depth error, while multi-operation editing yields the lowest edited-region SSIM, reflecting the additional demands of changing spatial layout and coordinating multiple edits.

\begin{table}[htbp]
\centering
\caption{Supplementary operation-specific SSIM and depth metrics for JointEdit3D on SceneEdit3D-Bench.}
\label{tab:operation_specific}
\begingroup
\small
\setlength{\tabcolsep}{8pt}
\begin{tabular}{@{}l cc cc@{}}
\toprule
Operation & Edit SSIM$\uparrow$ & Bg. SSIM$\uparrow$ & Abs Rel$\downarrow$ & $\delta{<}1.25\uparrow$ \\
\midrule
Delete & 0.836 & 0.914 & 0.0540 & 0.9819 \\
Add & 0.698 & 0.921 & 0.0592 & 0.9741 \\
Move & 0.698 & 0.913 & 0.0858 & 0.9305 \\
Appearance & 0.840 & 0.926 & 0.0606 & 0.9719 \\
Multi-op & 0.672 & 0.892 & 0.0683 & 0.9628 \\
\bottomrule
\end{tabular}
\endgroup
\end{table}

\subsection{Region-wise Geometry and Confidence Filtering}
\label{app:region_geometry}

Regional geometry metrics in Table~\ref{tab:external_reconstruction} use renderer-defined edit and background regions with the same full-scene Sim(3) alignment. JointEdit3D lowers CD in both regions relative to Omni-3DEdit + VGGT, supporting structural editing together with scene preservation.

Confidence filtering controls the accuracy--completeness trade-off of the native point cloud. At a threshold of 0.85, filtering lowers prediction-to-reference error relative to the unfiltered output while reducing coverage. The filtered output outperforms \emph{JointEdit3D RGB + VGGT} on all five metrics reported in Table~\ref{tab:external_reconstruction}. The threshold was adjusted using renderer-referenced geometry scores, so the filtered row is marked as a post-hoc diagnostic ($^\dagger$) and excluded from ranking.

\subsection{Main-State Feedback Ablation}
\label{app:main_state_feedback}

The feedback ablation in Table~\ref{tab:ablation} tests the main-state-to-SceneAnchor feedback term in Equation~\ref{eq:anchor_update}. Introduce a fixed gate $b\in\{0,1\}$:
\begin{equation}
\mathbf{c}_j = \mathrm{AnchorBlock}_j\!\left(
\mathbf{c}_{j-1} + b\,W_{\mathrm{in}}^{(j)}(\mathbf{x}_{\ell_j}),\;
W_e^{(j)}(\mathbf{e}),\;W_{\mathrm{ctx}}^{(j)}(\mathbf{ctx})\right).
\end{equation}
The full model uses $b=1$; the feedback-disabled variant uses $b=0$ at every injected layer throughout both training and inference. Source RGB conditioning, timestep and image/text context, and the anchor-to-main residual updates in Equation~\ref{eq:anchor_hint} remain enabled.

Both configurations use the same pretrained checkpoint, corresponding parameter initialization, branch width and depth, injection layers, zero-initialized output projections, data order, random seeds, optimizer, batch size, losses, 2K-step schedule, and evaluation settings. Results are reported only in Table~\ref{tab:ablation}, alongside the existing 2K-step full-model reference.

\subsection{RGB versus RGB+Geometry SceneAnchor Input}
\label{app:rgb_geo_anchor}

Table~\ref{tab:rgb_geo_anchor} tests whether adding estimated source geometry improves SceneAnchor conditioning under the 2K-step protocol of Table~\ref{tab:ablation}. The full model uses \([\mathbf{z}^{\mathrm{rgb}}_{\mathrm{src}};\,0]\); the alternative fills the geometry half with an estimated source geometry latent, \(\hat{\mathbf{z}}^{\mathrm{geo}}_{\mathrm{src}}\). Both variants retain joint RGB-geometry generation, so this experiment examines source conditioning rather than removal of the generated geometry stream.

RGB-only conditioning achieves better edit fidelity and geometry quality, with comparable background preservation. Explicit source geometry may reinforce the original structure in regions that need to change, making reference-guided structural edits harder. These results support using RGB observations for source-scene anchoring while allowing the joint generator to predict the edited geometry.

\begin{table}[htbp]
\centering
\caption{SceneAnchor input ablation under the 2K-step protocol of Table~\ref{tab:ablation}.}
\label{tab:rgb_geo_anchor}
\begingroup
\small
\setlength{\tabcolsep}{8pt}
\renewcommand{\arraystretch}{1.08}
\begin{tabular}{@{}l ccc@{}}
\toprule
SceneAnchor input & Edit LPIPS$\downarrow$ & Bg.\ LPIPS$\downarrow$ & CD$\downarrow$ \\
\midrule
$[\mathbf{z}^{\mathrm{rgb}}_{\mathrm{src}};\,0]$ & \bestnum{0.209} & \bestnum{0.103} & \bestnum{0.1053} \\
$[\mathbf{z}^{\mathrm{rgb}}_{\mathrm{src}};\,\hat{\mathbf{z}}^{\mathrm{geo}}_{\mathrm{src}}]$ & 0.284 & 0.105 & 0.1124 \\
\bottomrule
\end{tabular}
\endgroup
\end{table}

\subsection{Source-Observation Robustness}
\label{app:source_corruption}

We corrupt only SceneEdit3D-Bench source videos (motion blur, sensor-like noise, JPEG). The edited reference and the evaluation target stay clean, so Table~\ref{tab:source_corruption} measures the editor's sensitivity to source degradation with a clean reference.

\begin{table}[htbp]
\centering
\caption{Native geometry under source-only corruption on SceneEdit3D-Bench.}
\label{tab:source_corruption}
\begingroup
\small
\setlength{\tabcolsep}{8pt}
\renewcommand{\arraystretch}{1.08}
\begin{tabular}{@{}l cccc@{}}
\toprule
Source input & Acc.$\downarrow$ & Comp.$\downarrow$ & CD$\downarrow$ & F-score$\uparrow$ \\
\midrule
Clean & 0.1153 & 0.0886 & 0.1020 & 0.9702 \\
Motion blur & 0.1243 & 0.0891 & 0.1067 & 0.9627 \\
Sensor-like noise & 0.1218 & 0.0894 & 0.1056 & 0.9615 \\
JPEG artifacts & 0.1193 & 0.0884 & 0.1039 & 0.9649 \\
\bottomrule
\end{tabular}
\endgroup
\end{table}

JointEdit3D maintains stable geometry under the tested source corruptions, with modest changes in CD and F-score. Completeness remains nearly unchanged, indicating that source degradation primarily affects point accuracy rather than scene coverage.

\subsection{Runtime and Memory Details}
\label{app:runtime_memory}

Table~\ref{tab:runtime_memory} provides the runtime breakdown and peak memory for the 49-frame full-3D setting in Figure~\ref{fig:runtime_efficiency}. Inference runs on a single NVIDIA H200 GPU. RGB/video baselines include VGGT reconstruction as post-processing; optimization methods include per-scene fitting.

\begin{table}[htbp]
\centering
\caption{Runtime and memory details for the 49-frame full-3D efficiency comparison.}
\label{tab:runtime_memory}
\begingroup
\scriptsize
\setlength{\tabcolsep}{4pt}
\renewcommand{\arraystretch}{1.04}
\begin{tabular}{@{}l r r r r r@{}}
\toprule
Method & Time (s) & Gen./Opt. (s) & Post. (s) & Peak alloc. (GB) & Peak reserved (GB) \\
\midrule
JointEdit3D (Ours) & 11.94 & 11.94 & 0.00 & 29.07 & 50.33 \\
MVInpainter~\citep{mvinpainter} & 43.25 & 19.45 & 23.81 & 43.80 & 66.88 \\
Omni-3DEdit~\citep{omni3dedit} & 237.15 & 213.34 & 23.81 & 56.40 & 73.08 \\
SPIn-NeRF~\citep{spinnerf} & 1619.00 & 1619.00 & 0.00 & -- & -- \\
Gaussian Grouping~\citep{gaussiangrouping} & 486.00 & 486.00 & 0.00 & -- & -- \\
GaussianEditor~\citep{gaussianeditor} & 285.07 & 285.07 & 0.00 & -- & -- \\
\bottomrule
\end{tabular}
\endgroup
\end{table}

JointEdit3D uses five denoising steps followed by RGB and geometry decoding in one inference run, with no per-scene optimization or VGGT reconstruction post-process. Memory is blank where the timing logs did not record peak allocation.

\appendixdatasetdetails

\section{Evaluation Protocol Details}
\label{app:evaluation_protocol}

\subsection{360-USID and Real-Scene Quantitative Protocol}
\label{app:real_scene_protocol}

We evaluate real-scene object removal on 360-USID~\citep{aurafusion360}. Before/after 3DGS reconstructions are used solely to obtain a shared source trajectory. The edited reference is an object-free ground-truth frame from the dataset; SAM3-propagated masks provide the inputs required by mask-guided baselines.

Metrics are computed at poses with official after-removal images, using those real images as targets and the official evaluation masks for every method. Table~\ref{tab:delete_results} reports edit-region PSNR/LPIPS. The scoring targets are not 3DGS renders, and SAM3-propagated input masks are not used for scoring. Preprocessing failures are the only reason for frame exclusion.

\subsection{Inpaint360GS Protocol}
\label{app:inpaint360gs}

Table~\ref{tab:inpaint360gs} evaluates object removal on the 11-scene Inpaint360GS benchmark~\citep{inpaint360gs}. Inpaint360GS uses per-scene 3DGS optimization; JointEdit3D uses five denoising steps followed by RGB--geometry decoding. Shared evaluation conditions include the initial reconstruction, scene, poses, ground truth, and metrics, while each method retains its native inference path.

For JointEdit3D, we reconstruct the unedited scene with the official Inpaint360GS 3DGS pipeline and render a source video along a trajectory covering the official \texttt{inpaint\_camera} test poses. One selected object-free image serves as the editing reference; object-free images at the evaluation poses provide the scoring targets. We evaluate the generated RGB views directly at these poses using full-image and NBS (newly exposed background) metrics. The predicted point cloud remains the native geometry output, without conversion to 3DGS for scoring.

\begin{table}[htbp]
\centering
\caption{Object removal on the 11-scene Inpaint360GS benchmark. NBS denotes the newly exposed background region. Inpaint360GS is a per-scene optimization method; JointEdit3D and Omni-3DEdit are feed-forward.}
\label{tab:inpaint360gs}
\begingroup
\small
\setlength{\tabcolsep}{8pt}
\renewcommand{\arraystretch}{1.08}
\begin{tabular}{@{}l cc cc@{}}
\toprule
Method & PSNR$\uparrow$ & LPIPS$\downarrow$ & NBS PSNR$\uparrow$ & NBS LPIPS$\downarrow$ \\
\midrule
Inpaint360GS~\citep{inpaint360gs} & \secondnum{24.40} & \secondnum{0.1300} & 36.29 & 0.0078 \\
Omni-3DEdit~\citep{omni3dedit} & 22.38 & 0.1495 & \secondnum{36.77} & \secondnum{0.0072} \\
\textbf{JointEdit3D} & \bestnum{25.03} & \bestnum{0.1227} & \bestnum{37.24} & \bestnum{0.0066} \\
\bottomrule
\end{tabular}
\endgroup
\end{table}

JointEdit3D achieves the best full-image and newly exposed-region scores among the compared methods, demonstrating effective removal and background completion on this real-scene benchmark.

\subsection{Baseline Interfaces and Aggregation}
\label{app:baseline_protocol}

All methods are mapped to a common prediction layout and scored against fixed target frames and evaluation masks. Method-specific preprocessing follows the released interfaces summarized in Table~\ref{tab:baseline_interface_protocol}.

\begin{table}[htbp]
\centering
\caption{Baseline interfaces used in evaluation. Methods receive source observations and editing guidance according to their native interfaces; SEVA is a protocol-mismatched novel-view-synthesis model that does not consume the source video. Evaluation always uses the fixed benchmark masks and target images.}
\label{tab:baseline_interface_protocol}
\begingroup
\small
\setlength{\tabcolsep}{3pt}
\renewcommand{\arraystretch}{1.08}
\resizebox{\textwidth}{!}{%
\begin{tabular}{@{}>{\raggedright\arraybackslash}p{10.5em}>{\raggedright\arraybackslash}p{15.0em}c >{\raggedright\arraybackslash}p{12.5em}@{}}
\toprule
Method & Inputs used in evaluation & Per-scene fitting & Reporting scope \\
\midrule
SPIn-NeRF~\citep{spinnerf} & Edit masks, camera poses, per-scene NeRF assets & Yes & Shared deletion table only \\
Gaussian Grouping~\citep{gaussiangrouping} & Segmentation/grouping cues, camera poses, per-scene 3DGS assets & Yes & Shared deletion table only \\
GScream~\citep{gscream} & Method-required masks/poses and per-scene 3DGS assets & Yes & Shared deletion table only \\
GaussianEditor~\citep{gaussianeditor} & Method-required masks/text/poses and per-scene 3DGS assets & Yes & Shared deletion table only \\
SEVA~\citep{seva} & Edited reference frame + target cameras only; no source video & No & Five-operation table as protocol-mismatched NVS row \\
MVInpainter~\citep{mvinpainter} & Source RGB views, edited reference, edit masks & No & Deletion and five-operation comparisons \\
Omni-3DEdit~\citep{omni3dedit} & Source RGB views, edited reference, camera/pose inputs & No & Deletion and five-operation comparisons \\
JointEdit3D & Source RGB video, edited reference, optional text prompt & No & Deletion and five-operation comparisons \\
\bottomrule
\end{tabular}%
}
\endgroup
\end{table}

Table~\ref{tab:baseline_coverage} marks unsupported properties and operations with a dash. Table~\ref{tab:delete_results} compares methods on deletion; Table~\ref{tab:main_operation_results} covers methods evaluated on all five operations. Each operation column averages that operation's samples. The Average column averages all 100 SceneEdit3D-Bench samples, using operation counts 29/29/14/14/14 (Delete/Add/Move/Appearance/Multi-op). The edited reference view is excluded from PSNR/LPIPS scoring on SceneEdit3D-Bench.

\paragraph{Evaluation regions.}
On SceneEdit3D-Bench, Edit denotes the renderer-defined affected-object mask, and Background denotes its complement. Both regions are evaluated against the edited target images. The Background region can include secondary effects outside the object mask, such as changed shadows and reflections; its scores measure agreement with the edited target in that region.

\paragraph{Shared inputs.}
Every method that can consume them gets the same source video and the same edited reference. If a method recommends a frame budget, we report the better of that budget and the all-frame setting; otherwise we use official defaults and do not tune on SceneEdit3D-Bench. Mask-guided methods receive the edit masks they cannot run without, and pose-conditioned methods receive benchmark cameras; JointEdit3D uses neither. An operation is protocol-mismatched when producing an output would change the task rather than just the file format---for per-scene NeRF/3DGS editors on add/appearance, that would mean plugging in an extra 2D editor to invent missing target content, which we do not score. SEVA is the one NVS diagnostic: it expands the edited reference to target cameras and never sees the source video.

\paragraph{Edit3r and the reference-input protocol.}
Edit3r~\citep{edit3r} applies its own 2D image-editing pipeline to the input views before predicting edited 3D Gaussians. Our reference-guided quantitative comparisons instead provide an externally edited reference frame to methods that support this input. Using Edit3r's own editor would introduce a different source of edit guidance, while adapting its editing stage to enforce the supplied reference would modify the native pipeline. We therefore discuss Edit3r as related work and omit it from the shared-reference quantitative comparison.

\paragraph{MVInpainter on Move and Multi-op.}
MVInpainter is mask-based multi-view inpainting, not a native Move editor; we include it as Omni-3DEdit did, using MVInpainter-F, which supports limited reference-guided insertion. Both methods get one edited reference, and MVInpainter additionally gets a mask. For Move, the mask composes the old location with a hand-drawn rectangle at the new one, not a ground-truth silhouette; for Multi-op, the edited-object masks are passed together in one forward pass. The Move results in Figure~\ref{fig:qualitative} characterize this adaptation of a removal-oriented checkpoint to simultaneous removal and insertion with the required inputs provided.

\subsection{3D Metric Protocol}
\label{app:3d_metric_protocol}

\paragraph{Reference and prediction sources.}
Table~\ref{tab:external_reconstruction} follows the Gen3R point-cloud protocol~\citep{gen3r}. The reference cloud is unprojected from edited-branch depth using renderer intrinsics and camera poses, with the same fixed frame subset and valid-pixel filtering for every method. All RGB reconstruction rows use the same VGGT checkpoint, resolution/crop policy, and inference settings. \emph{Target RGB + VGGT} provides an oracle RGB-input reference whose geometry is estimated by VGGT; it is excluded from ranking. \emph{JointEdit3D RGB + VGGT} reconstructs our generated RGB, while \emph{JointEdit3D} decodes geometry directly from the predicted joint latent.

\paragraph{Alignment and sampling.}
Each prediction is aligned to the renderer reference using one Sim(3) Umeyama transform estimated from full-scene valid correspondences. After removal of non-finite points, each predicted and reference cloud is subsampled to at most 100K points with a fixed random seed; smaller clouds retain all valid points. All global metrics use these sampled sets. Region-wise comparisons and confidence filtering in Table~\ref{tab:external_reconstruction} reuse the full-scene transform.

\paragraph{Metric definitions.}
Accuracy is the mean prediction-to-reference nearest-neighbor distance; Completeness measures the reverse direction; CD is their mean. F-score is the harmonic mean of precision and recall at a distance threshold of 0.20 in renderer world units. We interpret this threshold-specific score alongside the unthresholded Accuracy, Completeness, and CD distances. The separate post-hoc confidence-filtering procedure is described in Appendix~\ref{app:region_geometry}.

\paragraph{Native output and task scope.}
\label{app:native_3d_output}
JointEdit3D produces edited RGB views and native geometry aligned with the supplied source trajectory, without requiring a pre-existing NeRF or 3DGS as model input. The geometry outputs include depth, point maps, camera predictions, and a scene point cloud. Our evaluation measures edited-view quality, source-structure preservation, and native geometry accuracy. Editing along another trajectory uses source observations from that trajectory in a new inference pass. Rendering arbitrary novel views requires an additional downstream representation, such as a radiance field or 3DGS, beyond the native outputs evaluated here.

\section{Limitations}
\label{app:limitations}

JointEdit3D relies on the edited reference to specify the desired change. Errors or inconsistencies in that reference may propagate across views and into the generated geometry. Improving robustness to imperfect reference edits is a direction for future work.

\end{document}